%% file: main.tex
\documentclass[11pt,a4paper]{article}

\usepackage[utf8]{inputenc}
\usepackage[T1]{fontenc}
\usepackage{amsmath}
\usepackage{amsfonts}
\usepackage{amssymb}
\usepackage{graphicx}
\usepackage{hyperref}
\usepackage{geometry}
\usepackage{natbib}
\usepackage{float}
\usepackage{booktabs}
\usepackage{multirow}
\usepackage{xcolor}
\usepackage{wrapfig}
\usepackage{caption}
\usepackage{algorithm}
\usepackage{algorithmic}
\hypersetup{
    colorlinks=true,
    linkcolor=blue,
    citecolor=blue,
    urlcolor=blue,
    pdftitle={A Tutorial on Dimensionless Learning: Geometric Interpretation and the Effect of Noise},
    pdfauthor={Zhengtao Jake Gan, Xiaoyu Xie}
}

\title{A Tutorial on Dimensionless Learning: \\ Geometric Interpretation and the Effect of Noise}

\author{
    Zhengtao Jake Gan\textsuperscript{1}\\
    \textsuperscript{1}School of Manufacturing Systems and Networks\\
    Arizona State University\\
    \texttt{zhengtao.gan@asu.edu}
    \and
    Xiaoyu Xie\textsuperscript{2}\\
    \textsuperscript{2}Independent Researcher\\
    \texttt{xiaoyuxie.vico@gmail.com}
}

\date{\today}

\begin{document}

\maketitle

\begin{abstract}
Dimensionless learning is a data-driven framework for discovering dimensionless numbers and scaling laws from experimental measurements. This tutorial introduces the method, explaining how it transforms experimental data into compact physical laws that reveal compact dimensional invariance between variables. The approach combines classical dimensional analysis with modern machine learning techniques. Starting from measurements of physical quantities, the method identifies the fundamental ways to combine variables into dimensionless groups, then uses neural networks to discover which combinations best predict the experimental output. A key innovation is a regularization technique that encourages the learned coefficients to take simple, interpretable values like integers or half-integers, making the discovered laws both accurate and physically meaningful. We systematically investigate how measurement noise and discrete sampling affect the discovery process, demonstrating that the regularization approach provides robustness to experimental uncertainties. The method successfully handles cases with single or multiple dimensionless numbers, revealing how different but equivalent representations can capture the same underlying physics. Despite recent progress, key challenges remain, including managing the computational cost of identifying multiple dimensionless groups, understanding the influence of data characteristics, automating the selection of relevant input variables, and developing user-friendly tools for experimentalists. This tutorial serves as both an educational resource and a practical guide for researchers seeking to apply dimensionless learning to their experimental data.
\end{abstract}

\section{Introduction}
\input{introduction}

\section{Results and Discussion}
\input{results_discussion}

\section{Methods}
\input{methods}

\section*{Acknowledgements}
Z.~Gan thanks the support of U.S. Army DEVCOM Analysis Center (DAC) under contract W911QX25D0003 and Arizona State University startup funding, and acknowledges fruitful discussions with Aleksey Kashtelyan and David E. Mortin from DAC and Bo Shen from New Jersey Institute of Technology. X.~Xie thanks his wife and daughter for their patience and understanding during many late-night meetings.

\bibliographystyle{unsrt}
\bibliography{references}

\end{document}

%% file: introduction.tex

All physical laws can be expressed as dimensionless relationships, elegant and compact forms that reveal the fundamental scaling properties of nature. In 1883, Reynolds studied fluid flow through pipes and realized that velocity and viscosity need not be treated separately. A single dimensionless number, now called the Reynolds number, captures the key transition between laminar and turbulent flow. This is the power of dimensionless numbers: they strip away the arbitrary scales of our measurements and expose the underlying physics. More than 1200 such numbers have been discovered across diverse fields, from the Reynolds~\cite{buckingham1914} and Froude~\cite{froude1870} numbers in fluid mechanics and transportation to the Péclet~\cite{peletier1974} and Fourier~\cite{incropera2011} numbers in heat transfer, the Womersley number~\cite{womersley1955} characterizing pulsatile blood flow in biology, the Damköhler number~\cite{damkohler1936} in chemical reactions and metabolic processes, and even the Sharpe ratio~\cite{sharpe1966} in finance. Yet despite this century old foundation, discovering new dimensionless numbers and their relationships from experimental data remains surprisingly difficult, especially when measurements are scarce, noisy, or expensive to obtain. What if we could teach machines to discover these elegant scaling laws automatically, even from limited and imperfect data?

Dimensional analysis, established by Buckingham~\cite{buckingham1914} through the $\pi$ theorem, provides the mathematical framework. Any physically meaningful relationship between dimensional variables can be expressed in terms of dimensionless combinations, reducing complexity and revealing fundamental scaling laws. Dimensionless numbers, being ratios of forces, energies, or mechanisms, provide scale invariance and enable cross scale experiments through similitude theory~\cite{barenblatt1996}. However, traditional dimensional analysis has well documented limitations. The dimensionless numbers derived from Buckingham's $\pi$ theorem are not unique. Any basis for the null space of the dimension matrix is equally valid mathematically, but the theorem fails to identify which dimensionless groups are dominant for a specific physical system. Moreover, dimensional analysis alone cannot reveal the mathematical relationship between dimensionless numbers. Establishing scaling laws typically requires transforming experimental measurements into dimensionless numbers and fitting high dimensional response surfaces, a process that is time consuming and heavily relies on domain expertise through trial and error.

The integration of dimensional analysis with machine learning has emerged as a promising direction to overcome these limitations. Early efforts include the SLAW algorithm by Mendez and Ordonez~\cite{mendez2009}, which combines dimensional analysis with multivariate linear regression to identify power law relationships. However, SLAW assumes power law forms between input and output variables that are invalid in many applications. Constantine, Rosario, and Iaccarino~\cite{constantine2016,constantine2017} connected active subspace methods to dimensional analysis, revealing that all physical laws are ridge functions. Their method is only applicable to idealized systems with negligible noise.

The modern era of dimensionless learning began with DimensionNet, introduced as part of the HiDeNN framework~\cite{saha2021} in 2021. This work first combined Buckingham $\pi$ theorem with neural networks. The breakthrough came in 2022 with dimensionless learning by Xie and Gan et al.~\cite{xie2022dimensionless}, which proposed a mechanistic data driven approach embedding dimensional invariance into a two level machine learning scheme. This framework automatically discovers both dominant dimensionless numbers and governing laws from scarce measurement data. The same year, Bakarji et al.~\cite{bakarji2022} introduced BuckiNet, focusing on dimensionally consistent learning with Buckingham Pi. Villar et al.~\cite{villar2023} proposed exact units equivariance in machine learning models. More recently, Yuan and Lozano-Durán~\cite{yuan2025} proposed an information theoretic approach, demonstrating how information based criteria can guide the discovery of dimensionless groups.

Despite these significant advances, current dimensionless learning methods face critical limitations. Most implementations handle only one or two dimensionless groups, yet many applications like fatigue testing require four or five or more~\cite{carpinteri2009}. The methods lack geometric interpretation of the dimensional reduction space, obscuring how physical constraints are embedded and how the learning process navigates the null space manifold. They require additional incorporation of domain knowledge to narrow the solution space. They lack comprehensive analysis of noise effects using controlled synthetic data. Finally, they lack user friendly interfaces, creating barriers for experimentalists with data but limited coding expertise.

This tutorial addresses these limitations through three key contributions. First, we present a comprehensive geometric interpretation that elucidates how the null space of the dimension matrix defines a manifold of valid dimensionless combinations and how the learning process navigates this constrained space. Second, we conduct a systematic investigation of noise effects using synthetic data with controlled noise levels across multiple noise models including additive, multiplicative, systematic bias, and mixed scenarios. This work quantifies robustness and provides practical recommendations for parameter selection, regularization strength, and data requirements. Third, we introduce an open sourced user friendly interface with intuitive workflows for data preprocessing, dimensional analysis, dimensional filtering, and optimization discovery. This interface is designed to make dimensionless learning accessible to experimentalists and domain experts who may lack deep expertise in machine learning or dimensional analysis but have valuable data and questions about underlying scaling laws.

%% file: results_discussion.tex

\subsection{Pipeline Overview}

The dimensionless learning pipeline transforms a table of experimental measurements into compact scaling laws. Imagine you have a spreadsheet with several columns of physical quantities you measured in your experiments, like power, speed, temperature, and so on. These are your inputs. You also have one column showing the quantity you care about, like the depth of a weld or the efficiency of a process. This is your output. The output is already dimensionless, meaning you've divided it by one or more input variables to remove units. The goal of dimensionless learning is simple: find the minimum number of dimensionless groups needed to best capture how your inputs relate to your output. Each dimensionless group is just a power law combination of your input variables, like power divided by speed squared, or viscosity divided by density times length.

\begin{wrapfigure}{r}{0.7\textwidth}
\centering
\includegraphics[width=0.68\textwidth]{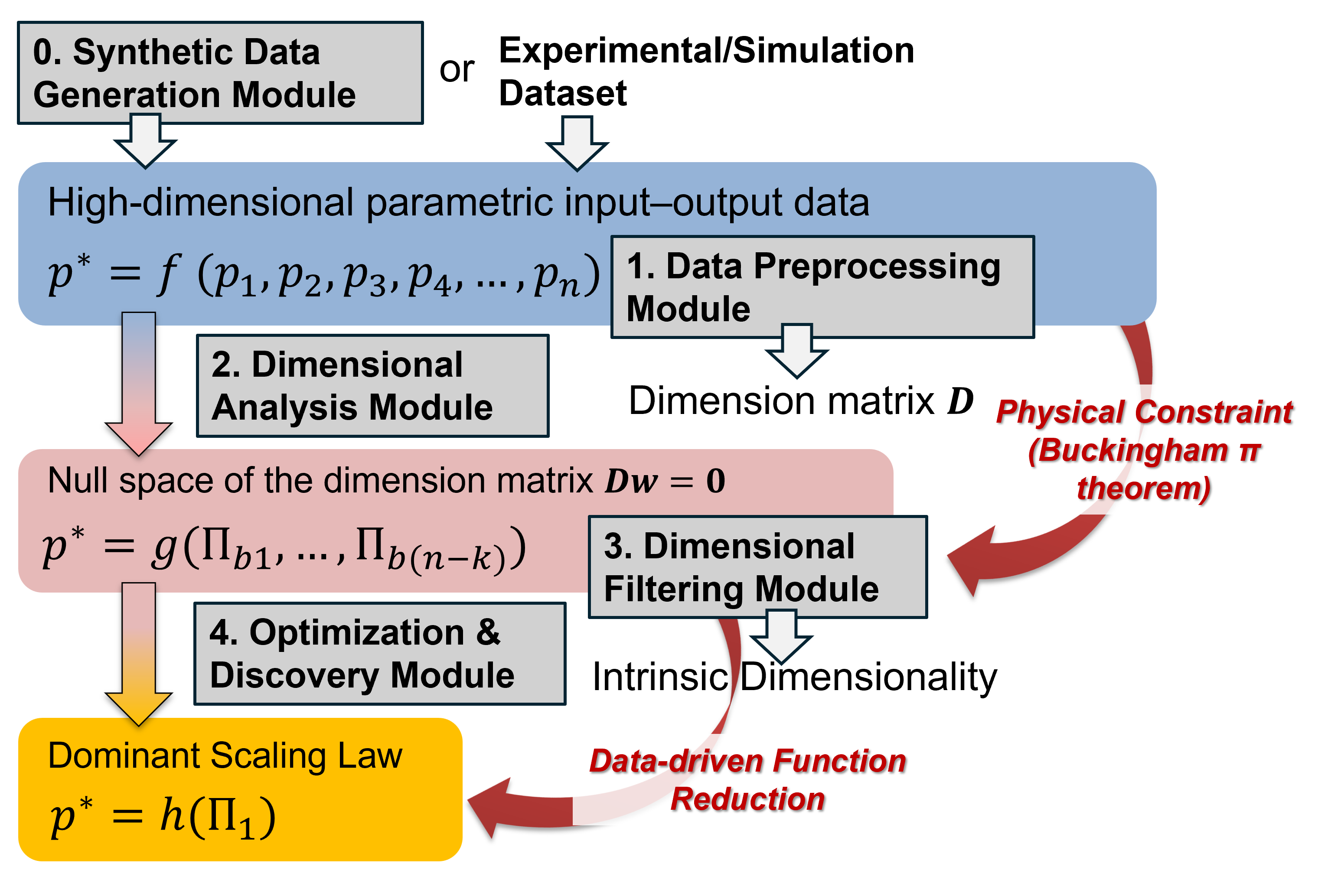}
\caption{Schematic diagram of the dimensionless learning pipeline workflow showing the five main modules that progressively transform data from high-dimensional inputs through basis space to compact scaling laws.}
\label{fig:pipeline}
\end{wrapfigure}

As illustrated in Figure~\ref{fig:pipeline}, the process unfolds through five main modules, each building on what came before. The workflow begins with either synthetic data generation or experimental datasets. Synthetic data generation creates test cases with known low-dimensional scaling patterns, where one or two dimensionless groups composed of inputs best represent the output. This validation step ensures the method works correctly before applying it to real experimental data.

Module 1 performs data preprocessing, where you select which columns are inputs and which is the output. The system maps your high dimensional input space, containing multiple physical quantities, down to a single output space. You identify the units and dimensions of each input variable, which gets encoded into a dimension matrix that tracks how each variable relates to fundamental units like mass, length, and time.

Module 2 conducts dimensional analysis by solving the null space structure of the dimension matrix. This finds all the different ways you can combine your input variables so that the units cancel out completely. The result is a set of basis vectors that span a reduced dimensional basis space. If you started with ten input variables, you might end up with six basis vectors, each representing one valid way to combine your inputs into something dimensionless.

Module 3 performs dimensional filtering to estimate how many of these basis vectors you actually need. Using techniques like principal component analysis, it looks at which directions in your data matter most for predicting the output. It suggests the dominant number of solutions in the null space that best predict the output, typically much fewer than the total number of basis vectors available.

Module 4 carries out optimization and discovery, searching for linear combinations of the basis vectors. Various optimization methods can be employed, including grid search, pattern search, or customized neural network architectures. In this tutorial, we demonstrate a neural network approach where the first layer searches for the best power coefficients of the input combinations, learning the gamma values that tell you exactly how to combine your inputs into optimal dimensionless groups. The network then applies a nonlinear mapping to the output, using a deep learning structure to capture complex but structured relationships between the dimensionless inputs and the output.

Module 5 delivers the final result: the form of the dimensionless number and the compact scaling law. The physical dimension reduction is complete. You've gone from a high dimensional input parameter space, say ten parameters, down through the null space basis, maybe six basis vectors, to just a few dominant dimensionless numbers, perhaps two or three, that map to your output. The scaling law shows you how these dimensionless groups relate to your quantity of interest, revealing the underlying physics in a simple, elegant form. The bottom panel of the figure visualizes this dimension reduction, showing how the high-dimensional input space transforms through the basis space to the final low-dimensional dimensionless space where the scaling law is discovered.

\subsection{Example Walkthrough: A Simple Case}

To make the pipeline concrete, let's walk through a simple example step by step. We'll generate synthetic data with seven input variables and one output, then follow it through all five modules to see how the method discovers the hidden scaling law.

We start with the data generation module. We create seven input variables $p_1, p_2, \ldots, p_7$, each with random values uniformly distributed between 0.5 and 1.0. These represent physical quantities we might measure in an experiment. The dimension matrix $\mathbf{D}$ encodes how each variable relates to fundamental units. For our example, we use four fundamental dimensions: mass, length, time, and temperature. The dimension matrix has seven columns, one for each input variable, and four rows for the fundamental dimensions. The actual dimension matrix is
\begin{equation}
\mathbf{D} = \begin{bmatrix}
2 & -2 & -1 & 2 & 2 & 2 & 0 \\
-2 & 0 & 0 & 1 & 1 & 0 & 0 \\
1 & -2 & 0 & -1 & 2 & -1 & 1 \\
0 & 2 & -2 & 1 & 2 & -2 & 2
\end{bmatrix}
\end{equation}
where each row corresponds to mass, length, time, and temperature respectively, and each column corresponds to $p_1$ through $p_7$.

The dimension matrix tells us that we have seven input variables but only four independent dimensions. This is exactly what Buckingham's $\pi$ theorem says: if you have $N$ variables and $r$ independent dimensions, you get $N-r$ dimensionless groups. But here's the beautiful part: if you've ever learned linear algebra, this becomes much clearer. The dimension matrix is just a matrix, and finding dimensionless groups is just finding the null space. If the matrix has full rank $r$, then the null space has $N-r$ basis vectors, corresponding to the number of independent dimensionless groups. We solve $\mathbf{D}\mathbf{w} = \mathbf{0}$, and the null space gives us exactly three basis vectors. The three basis vectors are
\begin{align}
\mathbf{w}_{b1} &= [0.259, 0.376, 0.803, 0.181, 0.337, 0, 0]^T \\
\mathbf{w}_{b2} &= [0.265, 0.059, 0.706, 0.530, 0, -0.383, 0]^T \\
\mathbf{w}_{b3} &= [0.177, 0.142, 0.781, 0.355, 0, 0, 0.461]^T
\end{align}
Each basis vector gives us one way to combine the inputs so that all units cancel out. We've normalized these basis vectors here, but you don't have to. Sometimes leaving them with integer components makes the analysis simpler later, especially when you want to interpret the physical meaning of each dimensionless group. These three basis vectors are our building blocks for creating dimensionless groups.

Now comes the next part. We assume that the target dimensionless number is a linear combination of these three basis vectors, with coefficients we'll call $\boldsymbol{\gamma}$. For our example, we set $\boldsymbol{\gamma} = [1, 1, 0]^T$. This means we take the first basis vector, add it to the second basis vector, and ignore the third. The final vector is computed as $\mathbf{w}_1 = \mathbf{w}_{b1} + \mathbf{w}_{b2} = [0.524, 0.435, 1.510, 0.711, 0.337, -0.383, 0]^T$. This tells us how to combine all seven inputs into a single dimensionless number using point-wise powers: $\Pi = \mathbf{p}^{\mathbf{w}_1}$, where the exponentiation is element-wise. The output is then computed using the compact scaling law
\begin{equation}
p^* = 2 + \Pi + 2\Pi^2
\end{equation}
We generate a hundred random data points this way, creating a complete dataset with inputs and outputs. The dataset and dimension matrix are saved for the next modules.

The data preprocessing module reads this dataset and the dimension matrix, but importantly, it doesn't know the hidden scaling pattern. It just sees seven inputs and one output. The module normalizes everything by dividing each variable by its maximum value, so all inputs and outputs end up below one. This normalization helps with numerical stability and makes the learning process easier. The module saves the normalized data and the dimension matrix for the next step.

The dimensional analysis module takes the dimension matrix and solves for the null space, giving us the same three basis vectors $\mathbf{w}_{b1}$, $\mathbf{w}_{b2}$, and $\mathbf{w}_{b3}$ we used in data generation. The module then constructs three basis dimensionless groups using element-wise powers:
\begin{align}
\Pi_{b1} &= \mathbf{p}^{\mathbf{w}_{b1}} = p_1^{0.259} p_2^{0.376} p_3^{0.803} p_4^{0.181} p_5^{0.337} \\
\Pi_{b2} &= \mathbf{p}^{\mathbf{w}_{b2}} = p_1^{0.265} p_2^{0.059} p_3^{0.706} p_4^{0.530} p_6^{-0.383} \\
\Pi_{b3} &= \mathbf{p}^{\mathbf{w}_{b3}} = p_1^{0.177} p_2^{0.142} p_3^{0.781} p_4^{0.355} p_7^{0.461}
\end{align}
These three dimensionless groups are the building blocks, but we don't know yet which combination we need.

Here's where the log transformation becomes crucial. Instead of working with the dimensionless groups directly, we take their logarithms. Why? Because raising inputs to powers and multiplying them becomes addition in log space. If we have $x^a y^b$, its logarithm is $a \log x + b \log y$. More generally, $\log(\mathbf{p}^{\mathbf{w}}) = \mathbf{w}^T \log(\mathbf{p})$, which transforms the power law into a linear combination. This is exactly what we need for the neural network. So we now have three log dimensionless groups $\log \Pi_{b1}$, $\log \Pi_{b2}$, $\log \Pi_{b3}$ plus the output $p^*$, all ready for the next module.

The dimensional filtering module analyzes this four dimensional dataset: three log dimensionless groups and one output. The question is: are these four dimensions all independent, or is there a hidden lower dimensional structure? In this case, $p^*$ is a function of $[\log \Pi_{b1}, \log \Pi_{b2}, \log \Pi_{b3}] \boldsymbol{\gamma}$, so we want to find the intrinsic dimension of the dataset after dimensional analysis. We use two methods (details in the Methods section). Principal component analysis (PCA) is accurate for linear data, gives an estimate for slightly nonlinear data, but fails for strongly nonlinear relationships. Sliced inverse regression (SIR) is a more robust method to find the intrinsic dimension in our case. In this simple example, both methods give high explained variance ratios for the first eigenvalue. PCA shows the first principal component explains 83.5\% of the variance, while the second explains only 7.8\%, the third 7.5\%, and the fourth 1.2\%. Sliced inverse regression gives an even stronger signal: the first direction explains 98.7\% of the variance, while the second explains only 0.8\% and the third 0.5\%. Both methods agree: we only need one independent dimensionless group, not three.

\begin{figure}[htbp]
\centering
\includegraphics[width=1.0\textwidth]{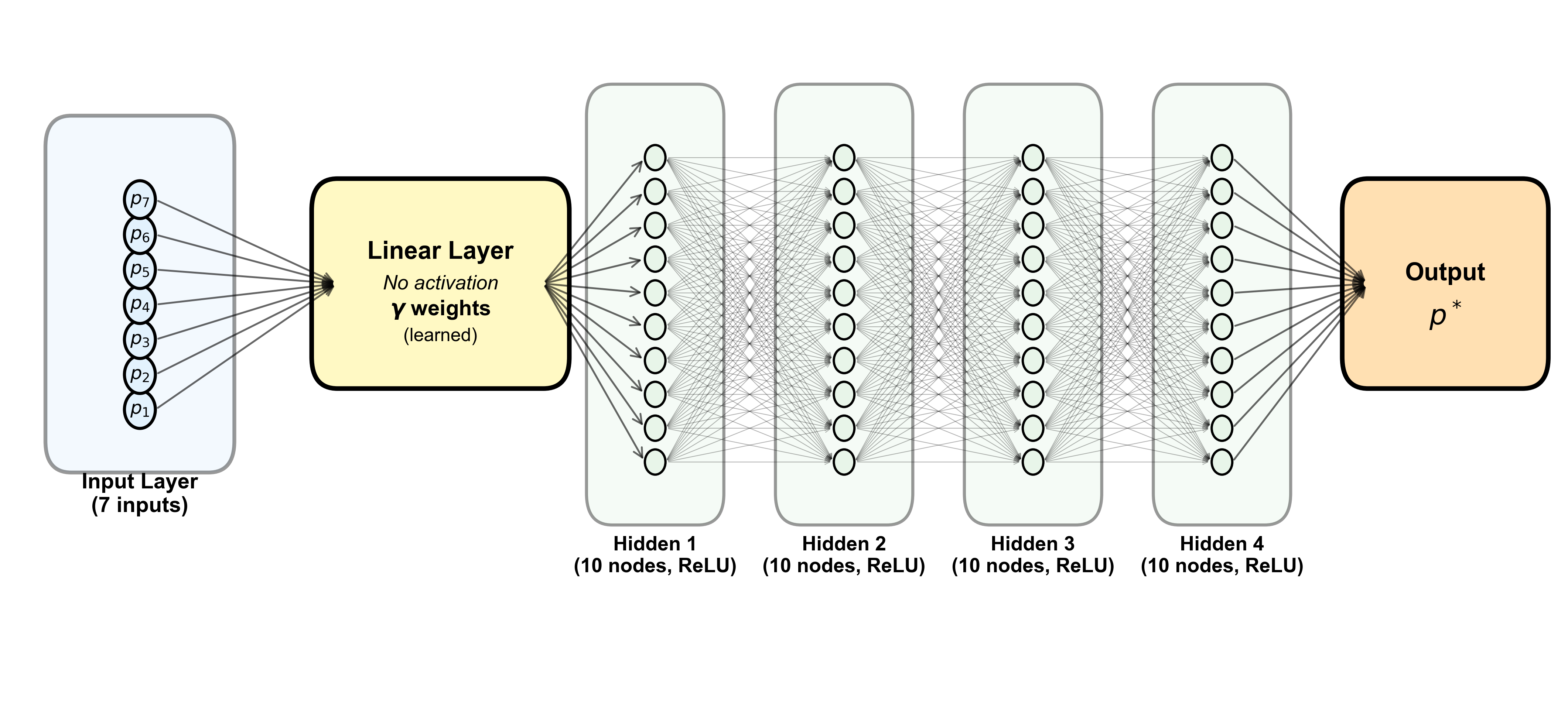}
\caption{Neural network architecture used in the example walkthrough, following the same structure as DimensionNet~\cite{saha2021}. The network has seven inputs, one linear combination layer (with no activation function) that learns the $\boldsymbol{\gamma}$ coefficients, four hidden layers with ten nodes each using ReLU activation, and one output. The linear layer weights directly represent the $\boldsymbol{\gamma}$ values that combine the basis dimensionless groups.}
\label{fig:nn_architecture}
\end{figure}

Armed with this knowledge, we set up the neural network in the optimization and discovery module. Figure~\ref{fig:nn_architecture} shows the architecture we use.

The network has seven inputs, one linear layer that learns the $\boldsymbol{\gamma}$ coefficients, four hidden layers with ten nodes each using ReLU activation, and one output. This architecture follows the same structure as DimensionNet~\cite{saha2021}, where the linear layer is special: it has no activation function, and its weights directly represent the $\boldsymbol{\gamma}$ values we're trying to learn. These weights tell us how to combine the three basis dimensionless groups into the final dimensionless number. The hidden layers then learn the nonlinear mapping from this dimensionless number to the output.

\begin{figure}[htbp]
\centering
\includegraphics[width=0.7\textwidth]{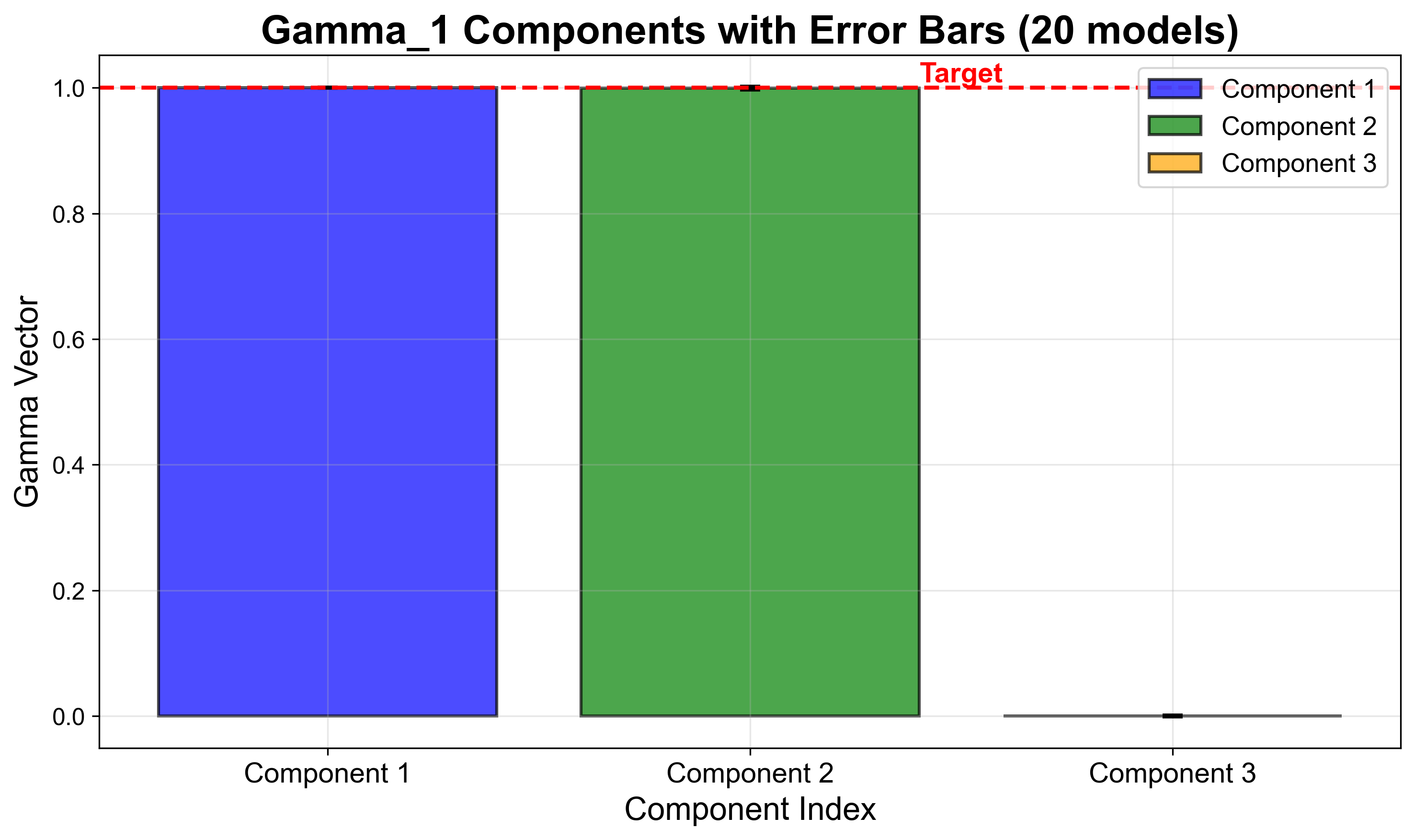}
\caption{Learned $\boldsymbol{\gamma}$ coefficients from twenty training runs with different random seeds, normalized so the first component is one. The first two components cluster around one, while the third clusters around zero, successfully recovering the target $\boldsymbol{\gamma}$ vector of $[1, 1, 0]^T$.}
\label{fig:gamma_case1}
\end{figure}

We train this network twenty times with different random seeds to capture the uncertainty in training. Each training run might converge to slightly different $\boldsymbol{\gamma}$ values, but they should all cluster around the true values if the method is working correctly. Figure~\ref{fig:gamma_case1} shows the results.

The plot displays the learned $\boldsymbol{\gamma}$ coefficients from all twenty training runs. But here's something interesting: we've normalized all these vectors so the first component equals one. Why? Because if we find that a dimensionless number with coefficients $\boldsymbol{\gamma} = [1, 1, 0]^T$ is dominant, then any scalar multiple of this vector, like $c\boldsymbol{\gamma} = [c, c, 0]^T$ for any constant $c$, will give us an equally dominant dimensionless number. The neural network doesn't know which representation we prefer, so it explores the whole one-dimensional subspace of equivalent coefficients. This is actually a beautiful feature of dimensional analysis: once we've identified a dominant dimensionless number, there's an entire linear subspace of equivalent coefficient vectors that all work equally well.

In traditional dimensional analysis, we manually prefer to use a dimensionless number with simple and integer coefficients, like $[1, 1, 0]^T$ rather than $[1.2, 1.2, 0]^T$ or $[0.75, 0.75, 0]^T$. But the neural network gives us the whole subspace. We normalize to the first component being one to make comparisons easier and to align with the traditional preference for simple representations. We'll see examples later that show how this linear subspace of equivalent representations works. The key point is: we can see that the first two components cluster around one, while the third component clusters around zero. This matches our target $\boldsymbol{\gamma}$ vector of $[1, 1, 0]^T$ remarkably well. The method successfully discovered the hidden dimensionless group.

\begin{figure}[htbp]
    \centering
    \includegraphics[width=1.0\textwidth]{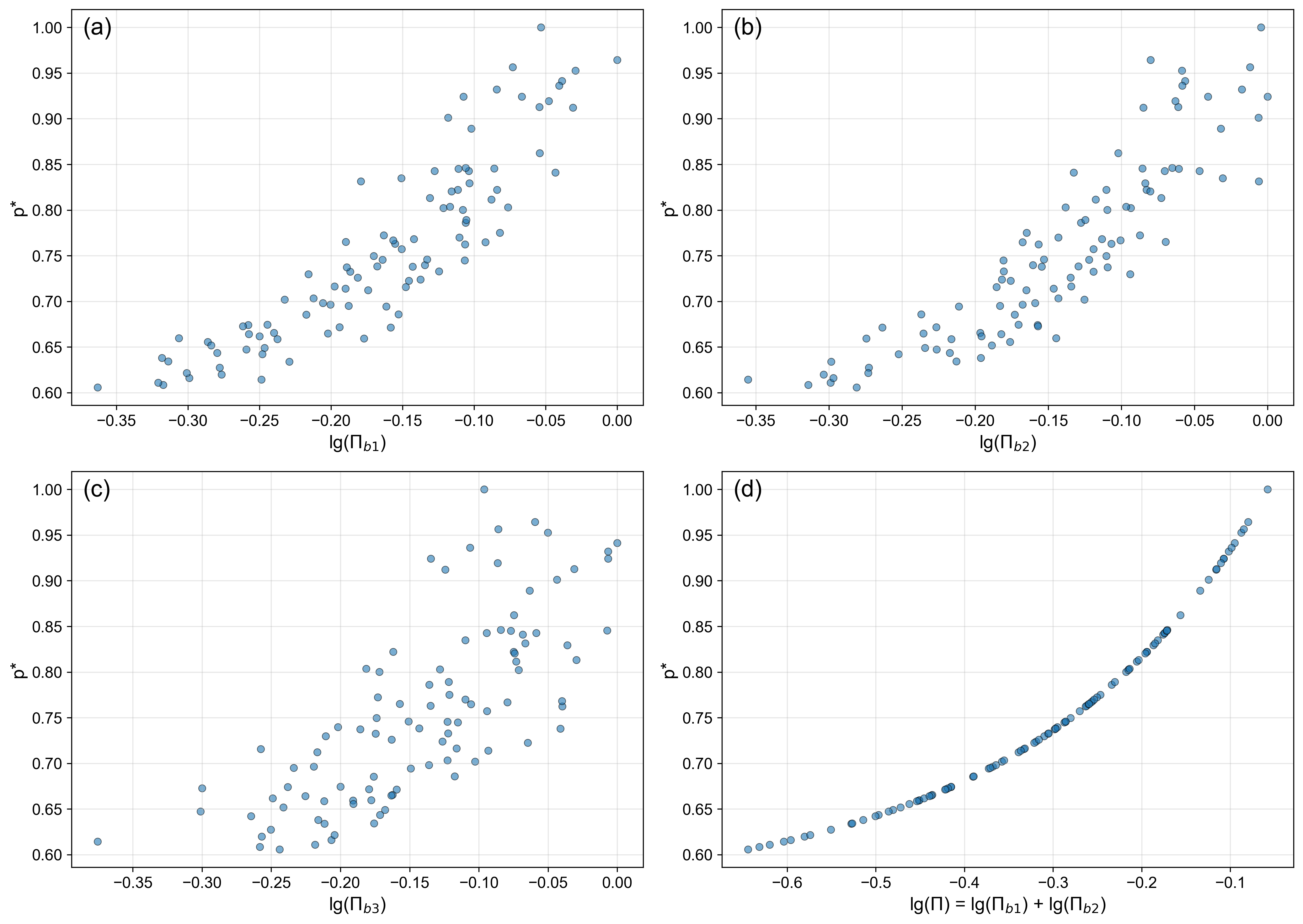}
    \caption{Correlation analysis between output and dimensionless groups. Subfigures (a), (b), and (c) show the correlations between the output $p^*$ and the three basis dimensionless groups $\log \Pi_{b1}$, $\log \Pi_{b2}$, and $\log \Pi_{b3}$ respectively. These plots reveal scattered relationships, indicating that no single basis dimensionless group alone captures the output structure. Subfigure (d) shows the discovered scaling law using the optimal combination $\log \Pi = \log \Pi_{b1} + \log \Pi_{b2}$, where the data points align with the polynomial relationship $p^* = 2 + \Pi + 2\Pi^2$, demonstrating successful dimension reduction from seven inputs to a single dimensionless group.}
    \label{fig:case1_data}
    \end{figure}

Finally, Figure~\ref{fig:case1_data} provides a clear visual demonstration of why dimensional filtering is essential. Subfigures (a), (b), and (c) show the correlations between the output $p^*$ and each of the three basis dimensionless groups $\log \Pi_{b1}$, $\log \Pi_{b2}$, and $\log \Pi_{b3}$ individually. As we can see, these relationships are quite scattered, with no clear pattern emerging from any single basis dimensionless group. This confirms what the dimensional filtering module told us: we cannot predict the output using just one basis dimensionless group alone.

The breakthrough comes in subfigure (d), which shows the discovered scaling law using the optimal linear combination $\log \Pi = \log \Pi_{b1} + \log \Pi_{b2}$ identified by the neural network. Here, the data points fall beautifully along a smooth curve, which is the polynomial relationship $p^* = 2 + \Pi + 2\Pi^2$ we encoded in the data generation. The method successfully reduced our seven dimensional input space down to a single dimensionless number that captures all the essential physics. What started as a complex problem with seven variables has been reduced to a simple, elegant scaling law that reveals the underlying physical relationship.

\subsection{Quantization Regularizer}

During the optimization process, the neural network learns the $\boldsymbol{\gamma}$ coefficients that tell us how to combine our basis dimensionless groups into optimal scaling laws. But here's the thing: if we don't put any constraints on the learning, the network can pick any numbers it wants. It might discover that the best coefficients are something like $[0.987, 1.234, -0.456]^T$, which works perfectly fine for making predictions, but it's not very pretty, is it? 

In traditional dimensional analysis, we like our dimensionless numbers to be clean and simple. We prefer coefficients that are integers or simple fractions, like 1, 2, or maybe 0.5, 1.5. These are much easier to interpret physically. So we introduce something called a quantization regularizer. It just adds a little penalty if the coefficients are far from simple values, encouraging the network to find solutions that are both accurate and interpretable. And here's a bonus: this regularization term also helps the method be more robust to noise by stabilizing the learned coefficients against small perturbations in the data. We'll see that in action in the next section.

The quantization regularizer is implemented as a soft penalty term added to the standard prediction loss. We use $\gamma_{ji}$ with two indices for multiple $\boldsymbol{\gamma}$ vectors. We might discover $k$ independent dimensionless groups, each needing its own $\boldsymbol{\gamma}$ vector ($\boldsymbol{\gamma}_1, \boldsymbol{\gamma}_2, \ldots, \boldsymbol{\gamma}_k$). The index $j$ identifies which group, and $i$ identifies which component of that group's vector. 

The double index $\gamma_{ji}$ is just a way to keep track of everything. The index $j$ tells you which dimensionless group you're looking at (the first one, second one, etc.), and the index $i$ tells you which component of that group's $\boldsymbol{\gamma}$ vector you're looking at. In our simple case study, we only have $k=1$ (one dimensionless group) and $N-r=3$ (three basis vectors), so we're really just working with a single $\boldsymbol{\gamma}$ vector with three components: $\boldsymbol{\gamma} = [\gamma_1, \gamma_2, \gamma_3]^T$. But the general formula works for any number of dimensionless groups.

Now, here's how the regularizer works: for each coefficient $\gamma_{ji}$, we look at all the simple target values we care about (like 0, 1, 2, or 0.5, 1.5, etc.) and find which one is closest. Then we measure how far away $\gamma_{ji}$ is from that nearest simple value. The regularization loss is just the sum of all these distances, weighted by a strength parameter $\lambda$:

\begin{equation}
\mathcal{L}_{\text{quant}} = \lambda \sum_{j=1}^{k} \sum_{i=1}^{N-r} \min_{s \in \mathcal{S}} |\gamma_{ji} - s|
\end{equation}

where $\lambda$ is the regularization strength (think of it as how much we care about simplicity versus fitting the data), $k$ is the number of dimensionless groups being learned, $N-r$ is the number of basis vectors (dimension of the null space), and $\mathcal{S}$ is the set of preferred simple values. The total loss function during training becomes:

\begin{equation}
\mathcal{L}_{\text{total}} = \mathcal{L}_{\text{prediction}} + \lambda \mathcal{L}_{\text{quant}}
\end{equation}

What simple values do we care about? Well, that depends on how simple you want to be. You can choose different resolutions. The simplest is just integers: 0, ±1, ±2, ±3, and so on. But we often use half integers, which gives us a bit more flexibility: 0, ±0.5, ±1.0, ±1.5, ±2.0, ±2.5, ±3.0. For our example, we use the half integer resolution, which means our target set is:
\begin{equation}
\mathcal{S}_{\text{half-integer}} = \{0, \pm 0.5, \pm 1.0, \pm 1.5, \pm 2.0, \pm 2.5, \pm 3.0\}
\end{equation}

Notice that zero is always included, no matter which resolution you choose. Why? Because zero is special: it means ``don't use this basis vector at all.'' This sparsity promoting property is really useful. In our example, we know the target is $\boldsymbol{\gamma} = [1, 1, 0]^T$, which means the third basis vector $\mathbf{w}_{b3}$ isn't needed. The regularizer will naturally push that third coefficient toward zero, helping the method automatically figure out which basis vectors are important and which ones aren't.

\begin{figure}[htbp]
    \centering
    \includegraphics[width=0.8\textwidth]{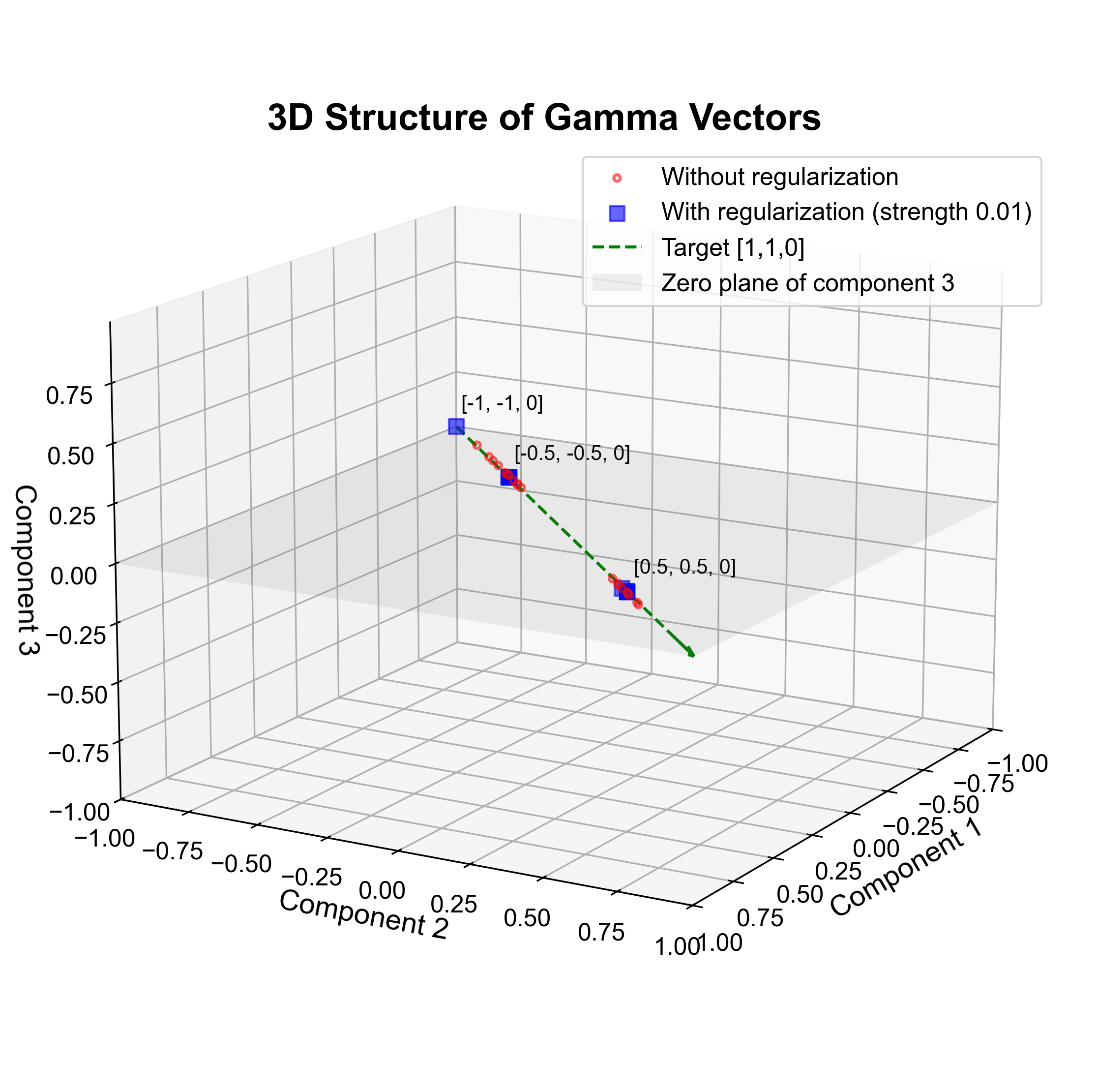}
    \caption{Three dimensional visualization of learned $\boldsymbol{\gamma}$ coefficient vectors from multiple training runs, showing the effect of quantization regularization. The plot displays the coefficients in the three dimensional space spanned by the basis vectors. The green dashed line represents the true direction $[1, 1, 0]^T$ (and all its scalar multiples $[c, c, 0]^T$), along which equivalent solutions lie. Without regularization, solutions are randomly distributed along this line. With quantization regularization, solutions cluster at three distinct half integer points: $[-1, -1, 0]^T$, $[-0.5, -0.5, 0]^T$, and $[0.5, 0.5, 0]^T$, making the discovered dimensionless groups more interpretable while maintaining predictive accuracy.}
    \label{fig:case1_3d_plot}
\end{figure}

Let's see what this looks like in action. Figure~\ref{fig:case1_3d_plot} shows us a three dimensional picture of where the learned $\boldsymbol{\gamma}$ vectors end up after training multiple times. Each axis represents one component of the $\boldsymbol{\gamma}$ vector, so we're looking at a three dimensional space where each point is a possible way to combine our three basis vectors.

Here's what's really interesting about this picture. The true answer $\boldsymbol{\gamma} = [1, 1, 0]^T$ doesn't just represent a single point in the three dimensional space: it defines a direction, because any scalar multiple $[c, c, 0]^T$ works equally well for the same dimensionless group. This means all equivalent solutions lie along a line in the three dimensional space, which is shown as the green dashed line in Figure~\ref{fig:case1_3d_plot}. 

Without regularization, the optimization finds solutions near this true direction, but they're randomly distributed along the line. The neural network might find coefficients like
\begin{equation*}
[0.987, 1.023, -0.015]^T \quad \text{or} \quad [-0.756, -0.743, 0.012]^T,
\end{equation*}
which work fine for fitting the data but are scattered and hard to interpret. With the quantization regularizer, the solutions are pushed toward specific clusters with simple half integer components. Instead of being randomly distributed along the line, the learned coefficients cluster at three distinct points: $[-1, -1, 0]^T$, $[-0.5, -0.5, 0]^T$, and $[0.5, 0.5, 0]^T$. These correspond to the nearest half integer values along the true direction, giving us interpretable solutions while maintaining the same predictive accuracy. The regularizer creates something like a Voronoi diagram in the coefficient space. This is especially useful when there are multiple equivalent solutions: remember, we talked earlier about how there's a whole subspace of $\boldsymbol{\gamma}$ vectors that all work equally well because they're just scalar multiples of each other. The regularizer breaks that symmetry by saying ``yes, they all work, but let's pick the simplest one.'' 

\subsection{Effect of Noise on Discovery}

So far, we've been working with clean synthetic data where everything is perfect. But in the real world, measurements are messy. The question is: does our method still work when the data is noisy? And if it does, how well does it work?

To answer this, we did a series of computer experiments. We took the same simple case we've been studying, with the true answer $\boldsymbol{\gamma} = [1, 1, 0]^T$, and we added noise to the data at different levels. Think of it like this: if the true value of something is 1.0, and you add 5\% noise, you might measure anywhere from 0.95 to 1.05. We tested noise levels from 0\% (perfect data) all the way up to 20\% (really noisy data). For each noise level, we used the same approach as before: we trained twenty different neural networks with different random seeds, creating an ensemble to capture the uncertainty. Each network learned its own $\boldsymbol{\gamma}$ vector, and we looked at how close they got to the true answer.

\begin{figure}[htbp]
    \centering
    \includegraphics[width=1\textwidth]{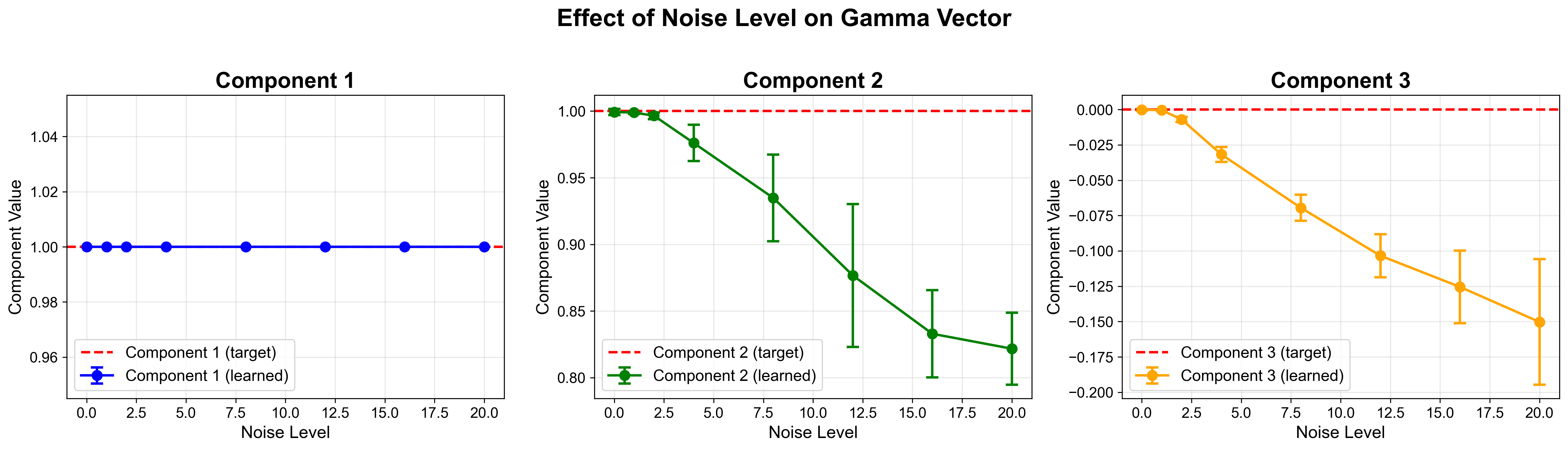}
    \caption{Effect of noise level on learned $\boldsymbol{\gamma}$ coefficients. Multiple test cases were generated with different noise levels in the data generation module, and the same neural network approach was used to learn the $\boldsymbol{\gamma}$ vectors. Each case was run with twenty ensembles to capture uncertainty (error bars shown). As noise increases, the learned coefficients drift away from the true values $[1, 1, 0]^T$, demonstrating the challenge of discovering dimensionless groups from noisy experimental data.}
    \label{fig:noise_effect_gamma}
\end{figure}

Figure~\ref{fig:noise_effect_gamma} shows what we found. The plot shows how the learned $\boldsymbol{\gamma}$ coefficients change as we add more and more noise to the data. You can see that as the noise level increases, the learned coefficients start to drift away from the true values. Without any noise, all the solutions cluster nicely around $[1, 1, 0]^T$, just like we saw before. But as you add noise, things start to scatter. At low noise levels (say, 1\% or 2\%), the solutions are still pretty close to the truth. But as you get to higher noise levels (10\%, 15\%, 20\%), the solutions spread out more and more. The method is still finding solutions that work for fitting the noisy data, but they're not necessarily the right dimensionless group anymore.

Now, here's something interesting. Before we even get to the optimization step, we need to figure out how many dimensionless groups we actually need. Remember, that's what the dimensional filtering module does: it looks at the data and tells us ``you probably need one dimensionless group'' or ``maybe two'' or ``definitely three.'' But what happens when the data is noisy? Does it still give us the right answer?

\begin{figure}[htbp]
    \centering
    \includegraphics[width=0.8\textwidth]{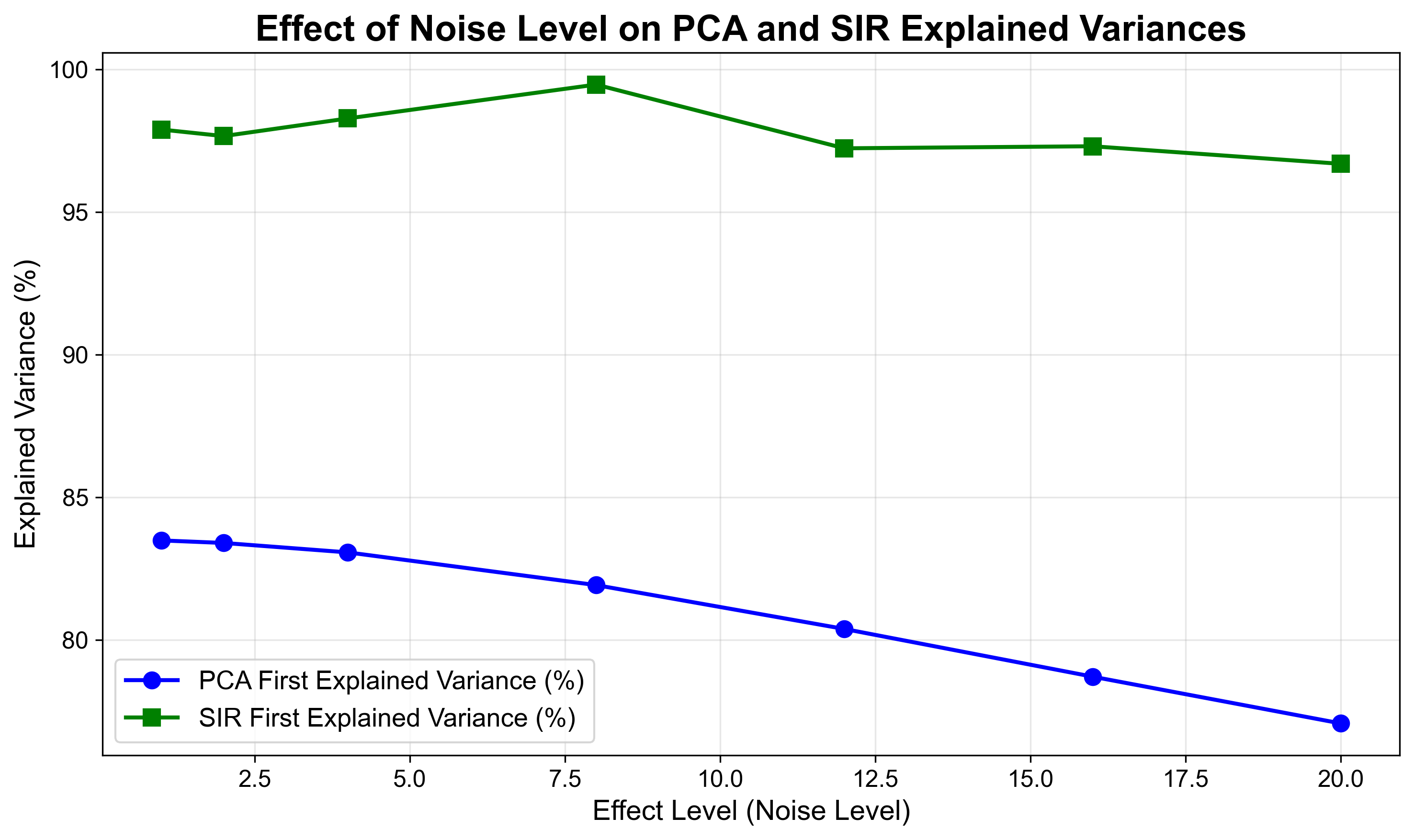}
    \caption{Effect of noise level on the first explained variance from PCA and SIR methods used in dimensional filtering. The plot shows how noise affects the estimation of intrinsic dimension. SIR demonstrates better resistance to noise, maintaining first explained variance above 75\% (indicating one dominant dimensionless number) even at high noise levels, while PCA's performance degrades more rapidly with increasing noise.}
    \label{fig:noise_effect_explained_variance}
\end{figure}

Figure~\ref{fig:noise_effect_explained_variance} shows how noise affects the dimensional filtering step. We look at two methods: PCA (principal component analysis) and SIR (sliced inverse regression). Both methods try to figure out the intrinsic dimension by looking at how much variance is explained by the first principal component or direction. If the first direction explains more than 75\% of the variance, that's a strong signal that we only need one dimensionless group. What you can see is that both methods work great when there's no noise: the first direction explains almost all the variance. But as noise increases, PCA starts to struggle. Its first explained variance drops pretty quickly. SIR, on the other hand, is much more robust. Even at 20\% noise, SIR still says ``the first direction explains more than 75\% of the variance, so you definitely need just one dimensionless group.'' That's really important because it means SIR can still give us the right answer about the intrinsic dimension even when the data is quite noisy.

\begin{figure}[htbp]
    \centering
    \includegraphics[width=1\textwidth]{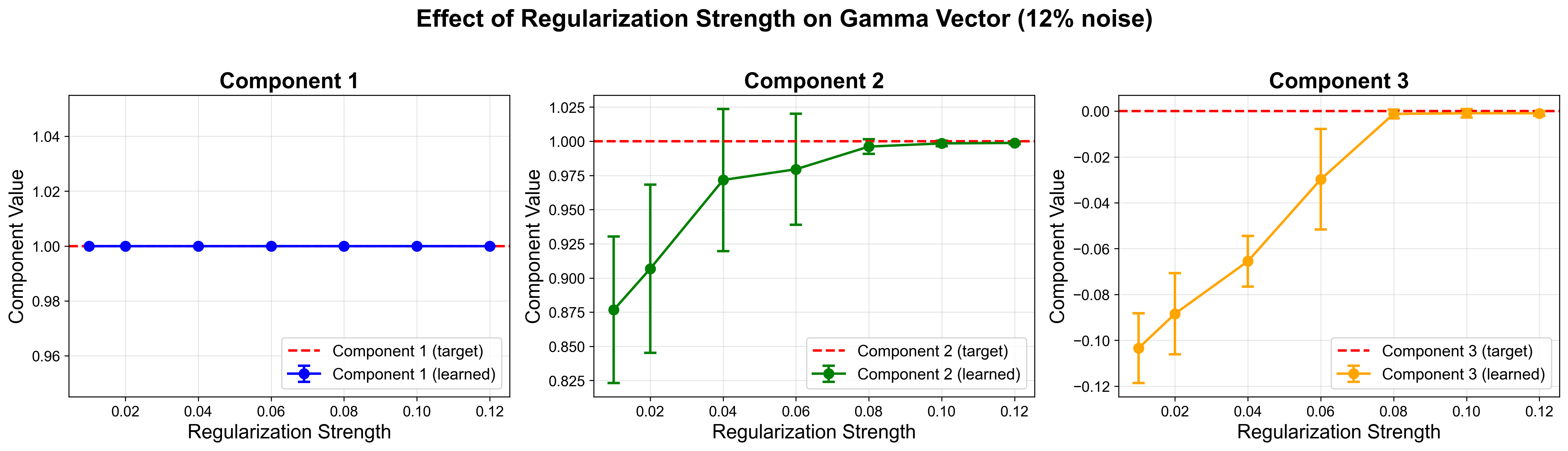}
    \caption{Effect of quantization regularization strength on learned $\boldsymbol{\gamma}$ coefficients at 12\% noise level. Multiple cases were tested with the same noise level (12\%) but different quantization regularization strengths ranging from 0 to 0.12. The figure clearly demonstrates that quantization regularization can suppress noise and produce accurate results even at noise levels representative of real experiments. As regularization strength increases, solutions cluster more tightly around the true values, showing the noise suppression mechanism in action.}
    \label{fig:gamma_reg_effect}
\end{figure}

Okay, so noise makes things harder. But remember that quantization regularizer we talked about earlier? We said it helps with noise resistance. Let's see if that's actually true. Figure~\ref{fig:gamma_reg_effect} shows what happens when we fix the noise level at 12\% (which is pretty representative of real experimental measurements) and vary the quantization regularization strength from 0 (no regularization) all the way up to 0.12 (strong regularization). The results are pretty clear. Without regularization ($\lambda = 0$), the solutions are all over the place. They're scattered randomly, and they're not very accurate. But as you increase the regularization strength, something beautiful happens: the solutions start to cluster around the right answer. At moderate regularization ($\lambda = 0.01$ or $0.05$), you start to see some clustering. And at strong regularization ($\lambda = 0.12$), the solutions cluster tightly around the true values $[1, 1, 0]^T$, or at least around the simple half integer values along that direction. The quantization regularizer is literally suppressing the noise. This is really important because 12\% noise is about what you'd expect in real experiments. So what we're seeing is that with the right amount of regularization, our method can still discover the correct dimensionless groups even from noisy experimental data. 

\begin{figure}[htbp]
    \centering
    \includegraphics[width=0.8\textwidth]{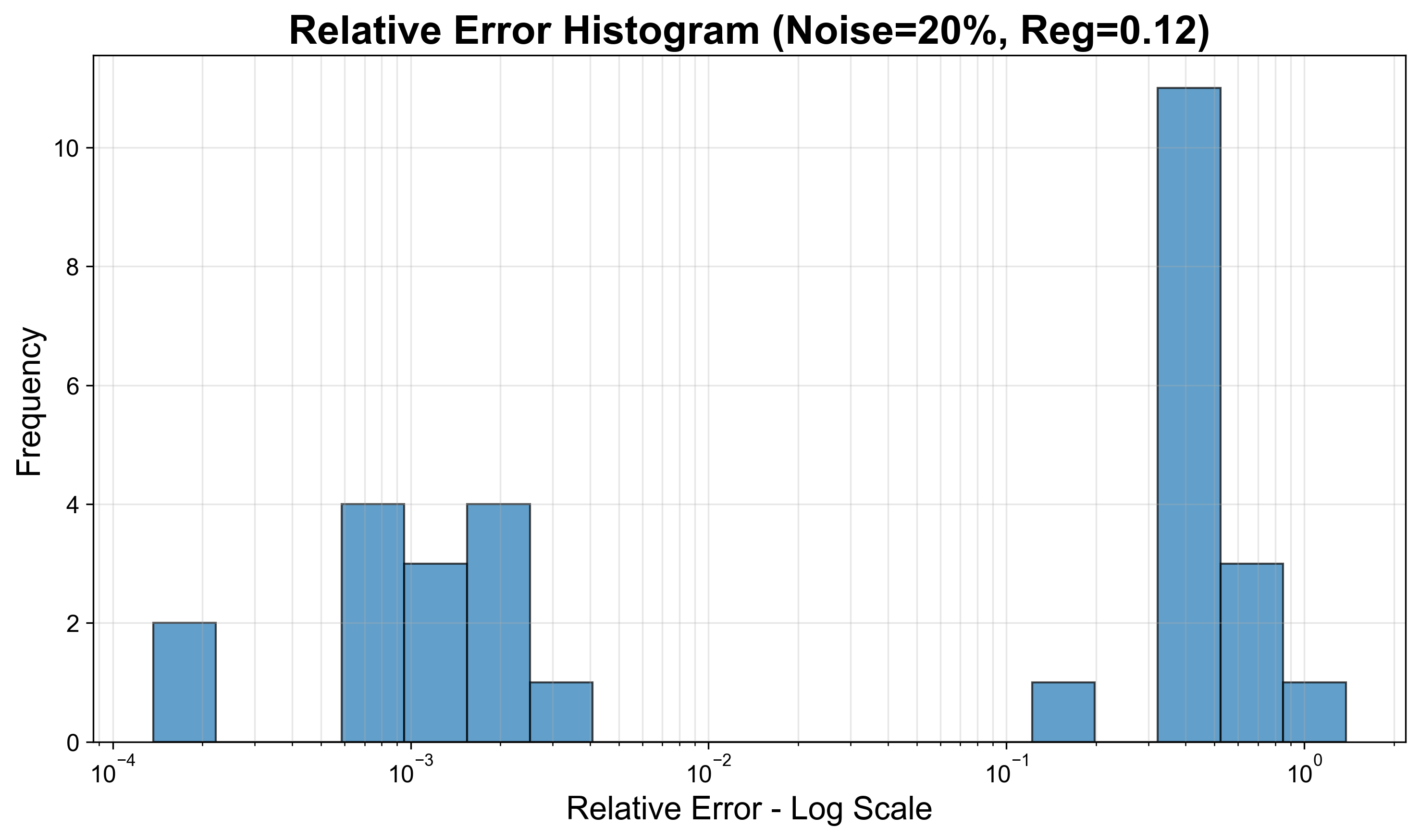}
    \caption{Histogram showing relative error between trained $\boldsymbol{\gamma}$ vectors and ground truth $[1, 1, 0]^T$ for twenty training runs with different random seeds, under extreme conditions of 20\% noise and high quantization regularization strength ($\lambda = 0.12$). At this extreme noise level, high regularization cannot always guarantee accurate solutions, but it pushes solutions into two distinct clusters: one with very accurate results (low relative error) and one with inaccurate results (high relative error). Solutions can be easily filtered by ranking $R^2$ error to identify the accurate cluster. This demonstrates the mechanism by which quantization regularization strength suppresses noise by creating clear separation between good and bad solutions.}
    \label{fig:case215_histogram}
\end{figure}

But what happens if you push things even further? What if the noise is really extreme, like 20\%? And what if you use really strong regularization? Figure~\ref{fig:case215_histogram} shows us a more extreme scenario: 20\% noise with regularization strength $\lambda = 0.12$. At this extreme noise level, even strong regularization can't always guarantee you'll get the right answer. But it does something interesting: it pushes the solutions into two distinct clusters. One cluster has very low relative error: these are the accurate solutions that got close to the true $[1, 1, 0]^T$. The other cluster has much higher relative error: these are the solutions that got pulled in the wrong direction by the noise. 

The histogram shows twenty different training runs, each with a different random seed. You can see that most of them fall into one of these two clusters. The key insight is that you can easily filter out the bad solutions by looking at their $R^2$ error, where $R^2$ (the coefficient of determination) measures how well the model predictions match the data, with values closer to 1 indicating better fit. The accurate solutions have low relative error and high $R^2$, while the inaccurate ones have high relative error and low $R^2$. So even when noise is extreme, the regularization doesn't give up: it creates a clear separation between good and bad solutions, making it easy to identify and keep only the accurate ones.

This figure also beautifully demonstrates the mechanism of how quantization regularization suppresses noise. When noise is strong, the data pulls you in random directions. But the regularizer creates these attractive basins at simple values. If the data pulls you close enough to a good simple value, the regularizer wins and you get an accurate solution. If the data pulls you too far away, you might end up at a different simple value (still interpretable, but not the right one). The regularization doesn't make noise disappear, but it organizes the solutions into clear clusters that we can easily filter.

There's another interesting challenge that comes up in real experiments. Think about material properties: density, thermal conductivity, viscosity, things like that. In theory, these are continuous variables. You could imagine having materials with density 1.0, 1.1, 1.2, 1.3, and so on, covering the whole range smoothly. But in practice, you're working with real materials. You might test steel, aluminum, copper, and a few other metals. Each material has its own discrete set of properties. You can't smoothly vary the density from 1.0 to 2.0: you have to jump from one material to another. So instead of having a continuous range of values, you end up with discrete samples.

\begin{figure}[htbp]
    \centering
    \includegraphics[width=1\textwidth]{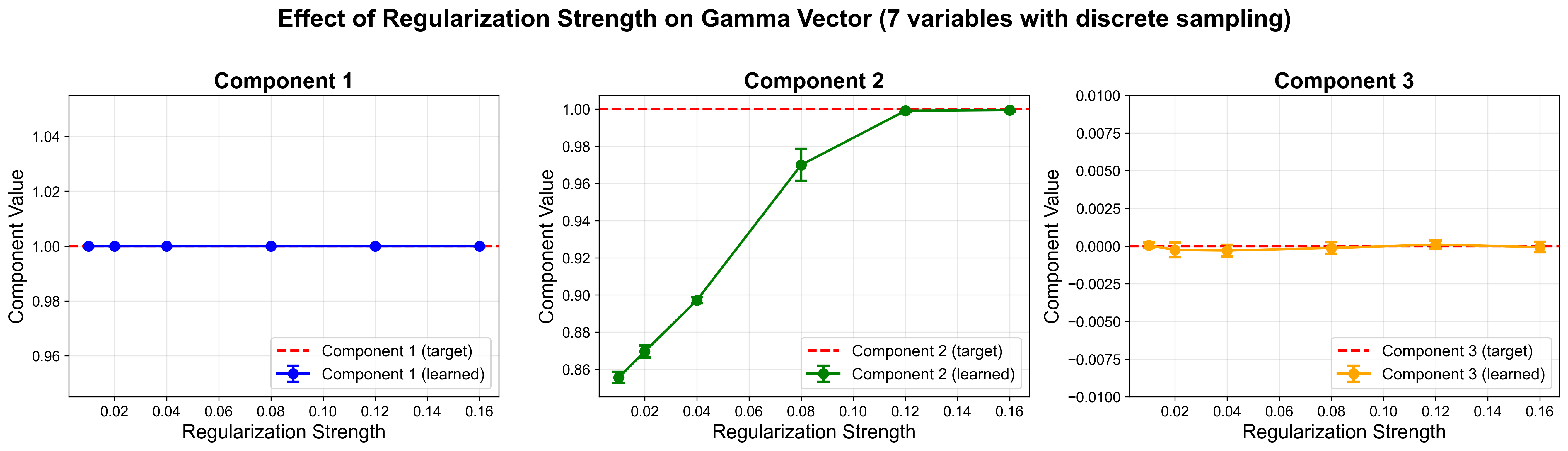}
    \caption{Effect of quantization regularization strength on learned $\boldsymbol{\gamma}$ coefficients for a case with seven discrete sampled input variables. Each variable has three random values sampled from 0.5 to 1.0, mimicking real experimental conditions where material properties are discrete rather than continuous. The figure shows how regularization strength (ranging from 0 to 0.16) helps suppress the negative influence of discrete sampling on dimensionless number discovery. As regularization strength increases, solutions cluster more tightly around the true values, demonstrating that quantization regularization is effective not only against noise but also against the challenges posed by discrete variable sampling.}
    \label{fig:gamma_case3_effect}
\end{figure}

The question is: does this discrete sampling affect our ability to discover dimensionless groups? To find out, we generated a test case with seven input variables, each sampled discretely. Instead of each variable taking on any value between 0.5 and 1.0, each variable was randomly assigned one of three discrete values in that range. So you might have variable 1 taking values \{0.6, 0.75, 0.9\}, variable 2 taking values \{0.55, 0.8, 0.95\}, and so on. This mimics the real experimental situation where you're testing different materials, each with its own fixed properties. What we found maybe not surprising: discrete sampling, just like noise, degrades the discovered $\boldsymbol{\gamma}$ vector away from the true value. The solutions start to scatter, and they're not as accurate. But here's the interesting part: quantization regularization can handle this situation too. Figure~\ref{fig:gamma_case3_effect} shows what happens when we vary the regularization strength from 0 (no regularization) all the way up to 0.16 (strong regularization) for this discrete sampling case. You can see the same pattern we saw with noise. Without regularization, the solutions are scattered all over the place. But as you increase the regularization strength, the solutions start clustering around the right answer. The regularizer is doing the same thing it did with noise: it's creating attractive basins at simple values, and those simple values happen to be close to the true dimensionless group. So even when your experimental data comes from discrete material samples rather than smoothly varying continuous variables, the quantization regularizer can still help you discover the correct dimensionless groups. This is really encouraging because it means our method works not just for idealized synthetic data, but for the messy, discrete reality of real experimental measurements.

\subsection{Multiple Dominant Dimensionless Numbers}

So far, we've been looking at cases where there's just one dominant dimensionless number. But what happens when the physics requires two or more independent dimensionless groups? This is actually quite common in real problems. Think about fluid dynamics: you might need both the Reynolds number and the Froude number to fully describe the flow. Or in heat transfer, you might need both the Nusselt number and the Prandtl number. The question is: can our method discover multiple dimensionless groups at once?

To test this, we generated a case with two dominant dimensionless numbers. We started with the same seven input variables and three basis vectors as before, but this time we created two independent dimensionless groups. The first one uses $\boldsymbol{\gamma}_1 = [1, 1, 0]^T$, which means we take the first basis vector plus the second basis vector. The second one uses $\boldsymbol{\gamma}_2 = [2, 0, 1]^T$, which means we take twice the first basis vector plus the third basis vector. The output is then computed as a function of both dimensionless groups: $p^* = \exp(2\Pi_1) + \Pi_2^{-0.5}$, where $\Pi_1$ and $\Pi_2$ are the two dimensionless numbers constructed from these $\boldsymbol{\gamma}$ vectors.

\begin{figure}[htbp]
    \centering
    \includegraphics[width=1.0\textwidth]{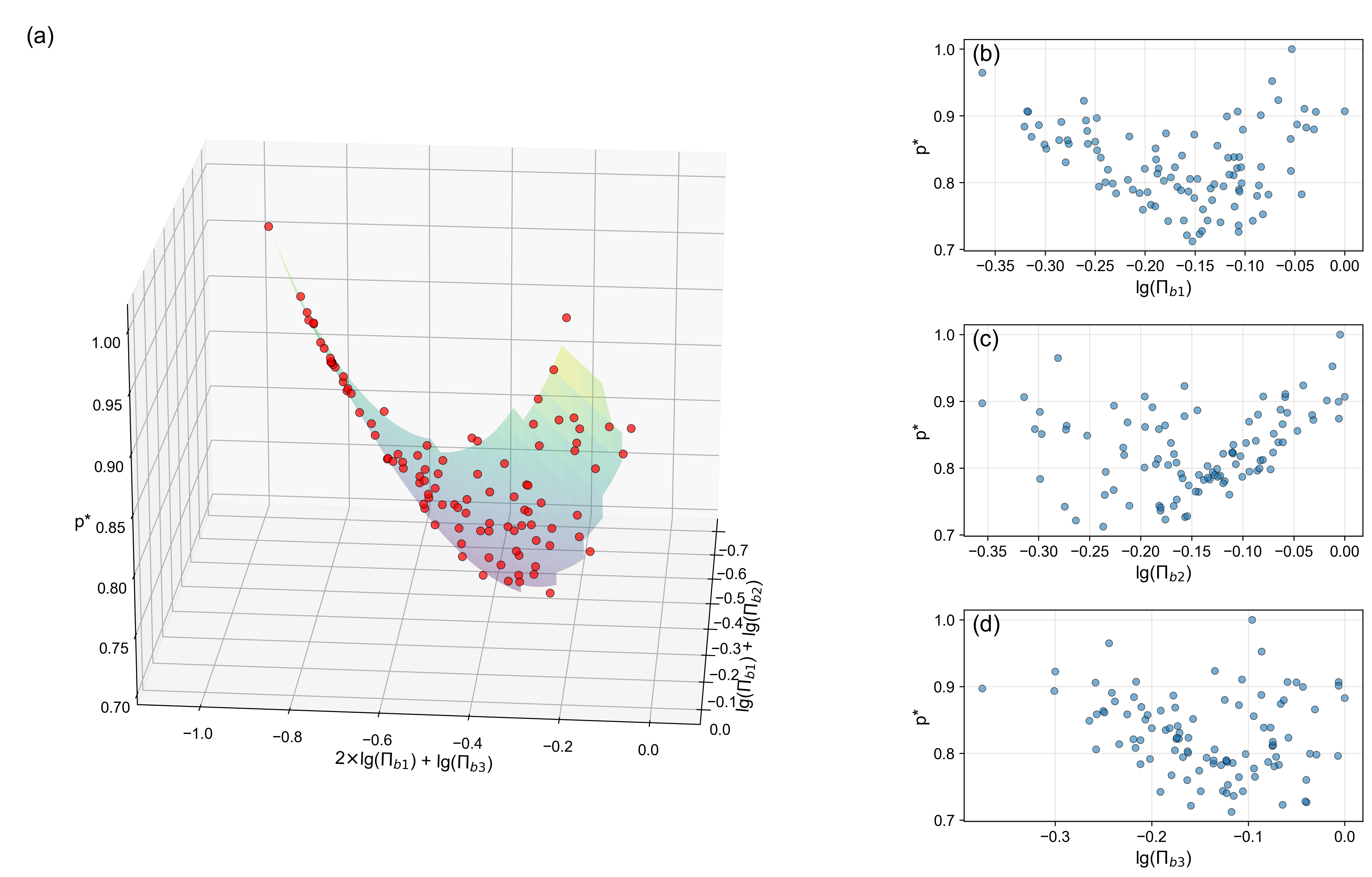}
    \caption{Correlation analysis for a case with two dominant dimensionless numbers. Subfigures (a), (b), and (c) show the correlations between the output and each of the three basis dimensionless groups individually, revealing scattered relationships that cannot capture the output structure alone. Subfigure (d) demonstrates the discovered two dimensional manifold in the data, coordinated by the two target dimensionless numbers $\Pi_1$ and $\Pi_2$, showing successful dimension reduction.}
    \label{fig:case4_data_plot}
\end{figure}

We generated a hundred data points with this structure, and ran the full pipeline. The dimensional filtering step correctly identified that we need two dimensionless groups. Then we trained the neural network with two linear combination layers, one for each dimensionless group. Figure~\ref{fig:case4_data_plot} shows what we discovered. The key observation is in how the data organizes itself. Subfigures (a), (b), and (c) show the correlations between the output and each of the three basis dimensionless groups individually. Just like in the single dimensionless number case, these are scattered. You can't predict the output using just one basis dimensionless group alone.

But look at subfigure (d). When we plot the output against the two discovered dimensionless groups $\Pi_1$ and $\Pi_2$, the data falls beautifully onto a two dimensional manifold. It's not a one dimensional curve anymore: it's a surface. The data organizes itself in a two dimensional space, and you can see the clear relationship $p^* = \exp(2\Pi_1) + \Pi_2^{-0.5}$ emerging from the scatter. This demonstrates that our method can successfully discover multiple dimensionless groups and reveal the multi dimensional structure hidden in the data.

Now, here's something really interesting. In the data generation, we assigned single point values for the $\boldsymbol{\gamma}$ vectors: $\boldsymbol{\gamma}_1 = [1, 1, 0]^T$ and $\boldsymbol{\gamma}_2 = [2, 0, 1]^T$. But when we look at what the neural network actually learned, something surprising happens.

\begin{wrapfigure}{r}{0.6\textwidth}
    \centering
    \includegraphics[width=0.58\textwidth]{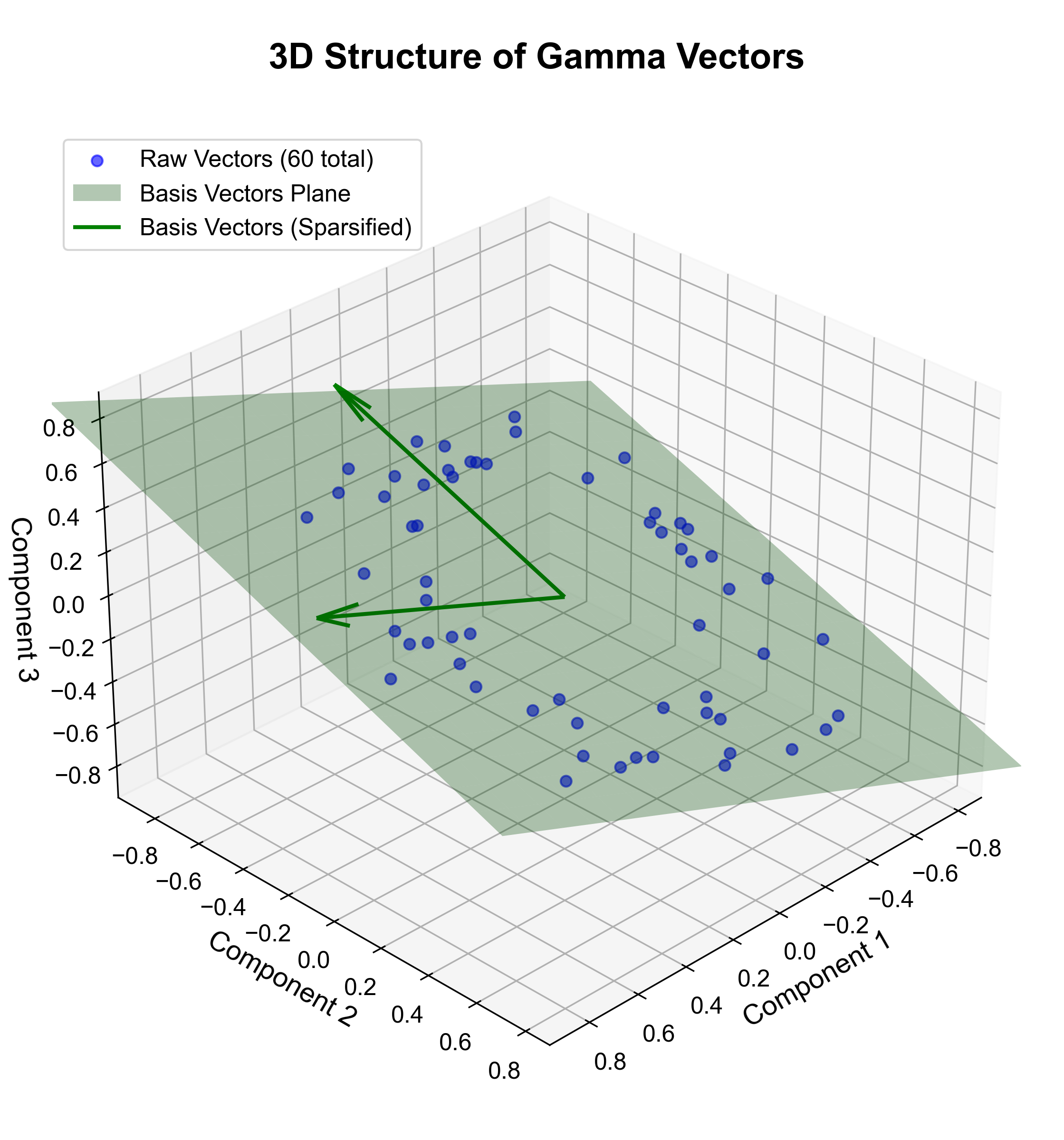}
    \caption{Three dimensional visualization of learned $\boldsymbol{\gamma}$ coefficient vectors for the two dimensional case.}
    \label{fig:case47_3d_plot}
\end{wrapfigure}

Figure~\ref{fig:case47_3d_plot} shows the learned $\boldsymbol{\gamma}$ vectors from multiple training runs, plotted in the three dimensional space of the basis vector coefficients. Instead of clustering around two points, the learned $\boldsymbol{\gamma}$ vectors form a plane! They're not randomly scattered: they lie on a well defined two dimensional surface in the three dimensional space. And here's the key insight: this plane is spanned by the two assigned $\boldsymbol{\gamma}$ vectors. This is mathematically profound.

Think about it this way. If the hidden scaling law is a function of two dimensionless groups, say $p^* = f(\log \Pi_1, \log \Pi_2)$, then mathematically, any function $f(x, y)$ can be rewritten as $g(ax + by, cx + dy)$ for any coefficients $a, b, c, d$. This means that if you have a scaling law that depends on $\log \Pi_1$ and $\log \Pi_2$, you can always rewrite it to depend on any linear combination of $\log \Pi_1$ and $\log \Pi_2$. So $f(\log \Pi_1, \log \Pi_2)$ is mathematically equivalent to $g(a\log \Pi_1 + b\log \Pi_2, c\log \Pi_1 + d\log \Pi_2)$, just with different functional complexity.

What this means for dimensionless learning is beautiful: we're not looking for a single set of dominant dimensionless numbers. Instead, there's an entire subspace of the gamma space that gives us mathematically equivalent representations. Any point in this subspace works equally well for the scaling law, but they differ in how simple or complex their dimensionless number expressions are. We can find the basis of this subspace and sparsify it to get the simplest form. In this case, the sparsified basis vectors are $[1.0, 0.0, 0.458]^T$ and $[0.0, -1.0, 0.499]^T$. The subspace spanned by these two sparsified basis vectors is very close to (essentially the same as) the subspace spanned by the assigned gamma vectors $[1, 1, 0]^T$ and $[2, 0, 1]^T$. In practice, an alternative approach is to search for a location in that subspace that gives us a simple form without resolving the whole subspace. Maybe we just find a point in the plane with simple dimensionless form, and that's good enough.

\begin{figure}[htbp]
    \centering
    \includegraphics[width=1.0\textwidth]{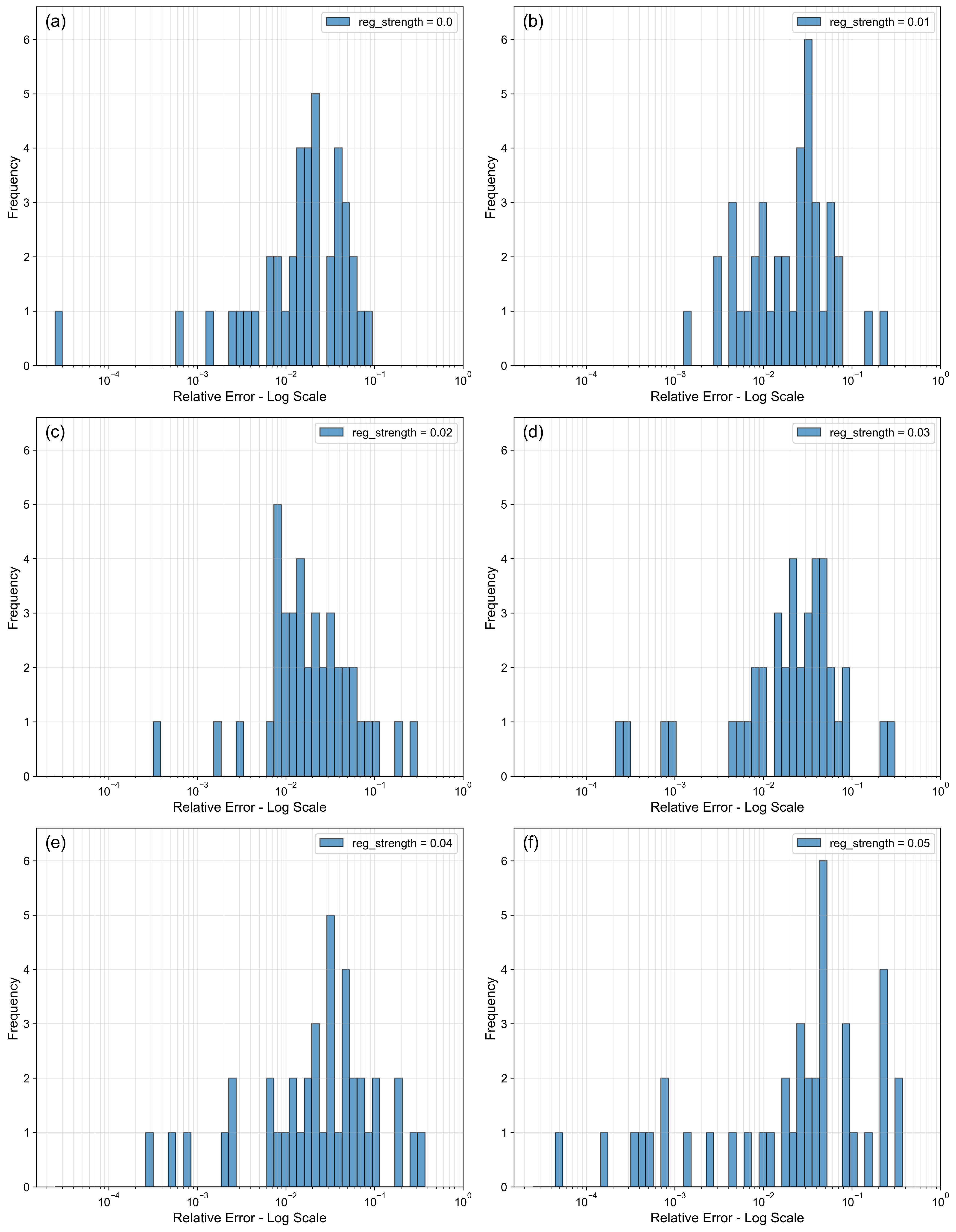}
    \caption{Relative error histograms for the two dimensional case with 5\% noise, showing the effect of quantization regularization strength. Subfigure (a) shows results without regularization, while subfigures (b) through (f) show increasing regularization strength. Thirty ensembles with different random seeds were used. Similar to the single dimensionless number case, quantization regularization pushes solutions toward discrete simple points, which spreads and improves the relative error distribution. Comparing subfigure (a) without regularization and subfigure (f) with regularization strength 0.05, the regularized case gives more accurate solutions and more diverse results with simple dimensionless forms. }
    \label{fig:gamma_projection_histograms}
\end{figure}

What about noise? Does quantization regularization still help when we have multiple dimensionless groups? To test this, we generated data with 5\% noise and ran the pipeline for thirty ensembles with different random seeds. Figure~\ref{fig:gamma_projection_histograms} shows the relative error histograms between the trained $\boldsymbol{\gamma}$ vectors and the true gamma vectors.

Subfigure (a) shows the results without regularization. The relative errors are distributed, but many solutions are not very accurate. As we increase the quantization regularization strength (subfigures (b) through (f)), something interesting happens. Just like in the single dimensionless number case, the regularization pushes solutions toward discrete simple points. But here's the key: this spreading of the relative error distribution is actually beneficial. Comparing subfigure (a) without regularization and subfigure (f) with regularization strength 0.05, the regularized case gives us two improvements. First, we get more accurate solutions: more training runs converge to solutions close to the true $\boldsymbol{\gamma}$ vectors. Second, we get more diverse results with simple dimensionless forms: the regularization encourages the discovery of different but equivalent simple representations within the subspace.

This tells us something important. At complex cases, whether it's high noise or multiple dimensionless numbers, it's very difficult to converge to the correct dimensionless numbers in a single attempt. But the quantization regularization provides a way to filter optimal solutions. By pushing toward simple values and creating clear clusters, it makes it easier to identify which solutions are accurate and which are not, even when the problem is complex.

\subsection{Interactive Web Interface}

\begin{figure}[htbp]
    \centering
    \includegraphics[width=1.0\textwidth]{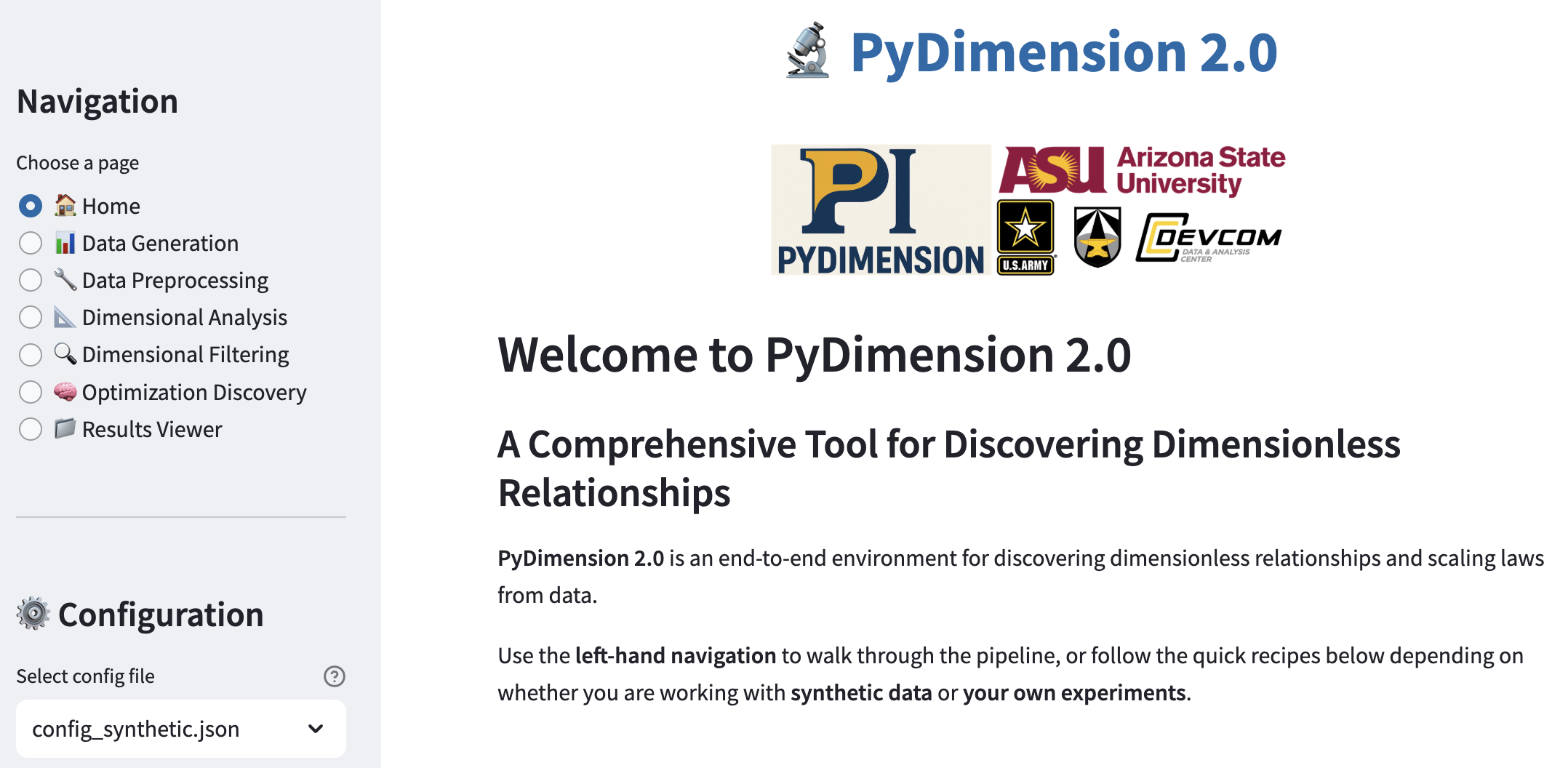}
    \caption{Interactive web interface for dimensionless learning. The homepage provides navigation to five sequential modules (Data Generation, Data Preprocessing, Dimensional Analysis, Dimensional Filtering, and Optimization Discovery) that guide users through the complete pipeline. Each module page includes configuration options, contextual guidance, and real-time visualizations.}
    \label{fig:website}
\end{figure}

To make dimensionless learning accessible to experimentalists and domain experts who may not have deep experience in machine learning or dimensional analysis, we developed an open-source web interface built on Streamlit (shown in Figure~\ref{fig:website}) and and publicly available at \url{https://huggingface.co/spaces/xiaoyuxie-vico/PyDimension}. The interface provides an intuitive, step-by-step workflow that mirrors the logical structure of the pipeline, allowing users to focus on their scientific questions rather than implementation details while still retaining the flexibility to adjust parameters when needed.

The interface organizes the pipeline into five sequential modules, each presented on its own dedicated page and aligned with the mathematical framework introduced earlier. Each module offers configuration panels and contextual explanations, helping users understand not only which parameters to set but why those choices matter for dimensionless learning.

\begin{itemize}
    \item \textbf{Data Generation} (optional): Creates synthetic datasets with known dimensionless relationships for testing and validation. Users can generate controlled benchmark datasets by specifying the number of variables, number of samples, and polynomial structures.

    \item \textbf{Data Preprocessing}: Manages data loading, variable selection, and normalization. Users may upload CSV files or select from pre-loaded datasets. The interface automatically detects input and output variables based on naming conventions but also supports manual adjustment. Interactive previews show statistical summaries, correlation matrices, and distribution plots to help users validate data quality.

    \item \textbf{Dimensional Analysis}: Computes basis vectors from the dimension matrix and constructs the corresponding dimensionless groups. Built-in guidance includes examples and reference tables for encoding physical dimensions (mass, length, time, temperature) as integer exponents. Users can visualize both the basis vectors and the resulting $\pi$-groups to understand how the original variables combine.

    \item \textbf{Dimensional Filtering}: Identifies the dominant dimensionless groups using PCA and SIR (other methods are under development). Users can adjust variance thresholds interactively to explore how the estimated intrinsic dimension changes. Explained variance ratios, eigenvalue spectra, and SIR direction plots are displayed to inform dimensionality reduction decisions.

    \item \textbf{Optimization Discovery}: Trains the neural network (or other regression methods) ensemble and extracts the discovered scaling laws. The interface automatically imports the suggested dominant dimension count from the filtering module as the default number of linear nodes. Training progress is monitored through loss curves and prediction comparisons. The discovered equations are displayed in both symbolic form and in terms of the original physical parameters, together with learned $\boldsymbol{\gamma}$ coefficients and performance metrics.
\end{itemize}

Several design principles guide the interface to reduce cognitive load and minimize user error. Progressive disclosure, implemented through expandable help sections and context-sensitive tooltips, provides layered guidance that supports new users without hindering experienced ones. Smart defaults and automatic parameter detection streamline the workflow by passing information between modules (for example, using the dominant dimension from filtering as the default number of linear nodes during optimization), while still allowing users to override any setting. Real-time feedback through progress indicators, inline visualizations, data previews, and diagnostic plots helps users quickly detect potential issues. Comprehensive input validation checks file structure, variable matching, and parameter ranges, producing clear error messages with actionable suggestions when problems arise.

The interface is implemented using Streamlit, a Python framework designed for rapid construction of interactive scientific applications. It directly calls the same \href{https://github.com/xiaoyuxie-vico/PyDimension}{PyDimension} modules used in command-line workflows, ensuring full consistency between web-based and programmatic usage. This architecture allows updates to the underlying modules to propagate automatically to both interfaces. Session state management preserves user configurations and intermediate results across pages, enabling iterative experimentation without losing previous progress. The complete source code is openly available at \url{https://github.com/xiaoyuxie-vico/PyDimension}, allowing users to inspect, modify, or extend the implementation as needed.

\subsection{Discussion}

The ultimate goal of dimensionless learning is to enable experimentalists to easily discover new dimensionless numbers with valuable physical insights from their data. Ideally, an experimentalist would be able to input their measurements and obtain clean, interpretable dimensionless numbers that reveal the underlying physics. However, this goal has not yet been fully realized.

The methods demonstrated in this work perform well on synthetic data and controlled test cases. We have shown successful discovery of dimensionless groups, robustness to noise, handling of discrete sampling, and identification of multiple dimensionless numbers. In the literature, the method has also been applied to conduct dimensionless learning for supersonic turbulence, aerodynamic drag, magnetohydrodynamic power generation, and laser-metal interaction~\cite{xie2022dimensionless,yuan2025}. However, there remains a significant gap between performance on controlled test cases and performance on real experimental data, where many challenges remain unsolved.

First, consider the challenge of multiple dimensionless numbers. While we have demonstrated successful discovery of two dimensionless groups, examples of recovering more than two dimensionless numbers from data are rare. The fundamental challenge is the exponential growth of the parametric search space with the number of dimensionless groups. Each additional dimension exponentially increases the computational complexity. More efficient methods are needed to search and identify the optimal subspace when dealing with many dimensionless groups. The curse of dimensionality and computational cost currently make it very challenging to extend the method beyond two or three groups.

Second, the effects of uncertainty and variation in real experimental data remain poorly understood. This tutorial has begun to examine noise level and discrete sampling, but many other factors require investigation. The number of sample points, the range of each variable, the distribution shape (normal, uniform, skewed, etc.), the sparsity of measurements, and correlations between variables may all affect the discovery process. Currently, we lack systematic understanding of how these factors interact and impact the identification of dimensionless groups. More comprehensive studies are needed to evaluate their effects and develop robust methods that can handle diverse data characteristics encountered in real experiments.

Third, input selection represents a critical problem that has not yet been addressed by any existing method. Currently, variable selection relies on human expertise, domain knowledge, and intuition. There is no systematic approach to determine which input variables should be included, which may lead to missing important variables or including irrelevant ones. However, this problem could potentially be quantified and automated through AI agents. An automated system could analyze correlations, test different variable combinations, and use information theory to identify the most relevant inputs. Such automation would significantly lower the barrier to using dimensionless learning, as users would no longer need dimensional analysis expertise to select appropriate variables.

Finally, there is the challenge of user accessibility. Currently, using dimensionless learning requires knowledge of the method, understanding of machine learning concepts, and programming skills. Most experimentalists simply want to analyze their data and extract physical insights without learning Python, understanding neural networks, or configuring regularization parameters. A user friendly interface is needed, potentially based on cloud computing or powered by large language models (LLM), where users can upload data, answer simple questions, and receive dimensionless numbers and scaling laws. Such an interface would make dimensionless learning accessible to users without expertise in dimensionless learning, machine learning, or coding.

In summary, significant progress has been made in dimensionless learning. The methods work effectively on synthetic data and controlled test cases, demonstrating successful discovery of dimensionless groups, robustness to noise, handling of discrete sampling, and identification of multiple dimensionless numbers. However, substantial challenges remain before dimensionless learning becomes a tool that any experimentalist can readily use. The fundamental challenge of multiple dimensionless numbers requires more efficient methods to search and identify the optimal subspace, as the exponential growth of the parametric search space makes it difficult to extend beyond two or three groups. The effects of uncertainty and variation in real experimental data need comprehensive investigation, extending beyond noise level and discrete sampling to include range, distribution shape, sparsity, and correlations between variables. Input selection remains a critical problem requiring automated methods that can reduce reliance on human expertise. Finally, user accessibility requires user friendly interfaces that make the method usable without knowledge of machine learning or coding. Addressing these challenges is essential for realizing the ultimate goal of making dimensionless learning accessible to experimentalists seeking physical insights from their data.

%% file: methods.tex

\subsection{Mathematical Framework}

Dimensionless learning formulates the discovery of scaling laws as an optimization problem. Given a dataset $\mathcal{D} = \{(\mathbf{p}_i, p^*_i)\}_{i=1}^{M}$ where $\mathbf{p}_i \in \mathbb{R}^N$ are input variables and $p^*_i \in \mathbb{R}$ is the output, the complete dimensionless learning framework can be expressed as a single comprehensive optimization problem:

\begin{equation}
\begin{aligned}
\min_{\{\boldsymbol{\gamma}_j\}_{j=1}^{k}, f} \quad & \underbrace{\frac{1}{M} \sum_{i=1}^{M} \left(p^*_i - f\left(\prod_{\ell=1}^{N} \tilde{p}_{\ell,i}^{\sum_{m=1}^{N-r} \gamma_{1m} w_{b,m\ell}}, \ldots, \prod_{\ell=1}^{N} \tilde{p}_{\ell,i}^{\sum_{m=1}^{N-r} \gamma_{km} w_{b,m\ell}}\right)\right)^2}_{\mathcal{L}_{\text{prediction}}} \\
& + \lambda \underbrace{\sum_{j=1}^{k} \sum_{i=1}^{N-r} \min_{s \in \mathcal{S}} |\gamma_{ji} - s|}_{\mathcal{L}_{\text{quant}}} \\
\text{subject to} \quad & \mathbf{D} \mathbf{w}_b^{(i)} = \mathbf{0}, \quad i = 1, \ldots, N-r \\
& \mathbf{w}_b^{(i)} \in \mathcal{N}(\mathbf{D}), \quad i = 1, \ldots, N-r \\
& k = \text{estimate}(\mathbf{X}, \mathbf{y})
\end{aligned}
\label{eq:complete_framework}
\end{equation}

This unified equation encapsulates the entire dimensionless learning pipeline. We now explain each component in detail:

\textbf{Objective Function Components:}

\begin{enumerate}
    \item \textbf{Prediction Loss $\mathcal{L}_{\text{prediction}}$}: The first term measures how well the learned scaling law predicts the output. It consists of:
    \begin{itemize}
        \item Normalized inputs $\tilde{p}_{\ell,i} = p_{\ell,i} / \max_j p_{\ell,j}$: Raw measurements scaled to be less than 1 for numerical stability.
        \item Basis weight vectors $\mathbf{w}_b^{(i)}$: Fixed vectors spanning the null space $\mathcal{N}(\mathbf{D})$ of the dimension matrix $\mathbf{D}$, ensuring dimensional homogeneity.
        \item Learnable coefficients $\boldsymbol{\gamma}_j = [\gamma_{j1}, \ldots, \gamma_{j(N-r)}]^T$: Parameters that linearly combine basis vectors to form dimensionless group weights $\mathbf{w}_j = \sum_{m=1}^{N-r} \gamma_{jm} \mathbf{w}_b^{(m)}$.
        \item Dimensionless groups $\Pi_{j,i} = \prod_{\ell=1}^{N} \tilde{p}_{\ell,i}^{\sum_{m=1}^{N-r} \gamma_{jm} w_{b,m\ell}}$: Products of powers of input variables, constructed from the learned weight vectors.
        \item Learned function $f: \mathbb{R}^k \rightarrow \mathbb{R}$: A neural network that maps the $k$ dimensionless groups to the predicted output $\hat{p}^*_i$.
        \item Mean squared error: Averages the squared prediction errors across all $M$ data samples.
    \end{itemize}
    
    \item \textbf{Quantization Regularization $\mathcal{L}_{\text{quant}}$}: The second term encourages interpretable coefficients by penalizing deviations from simple target values:
    \begin{itemize}
        \item Target set $\mathcal{S}$: Typically contains integers, half-integers, or quarter-integers (e.g., $\{0, \pm 0.5, \pm 1, \pm 1.5, \pm 2, \ldots\}$), always including zero to promote sparsity.
        \item Minimum distance penalty: For each coefficient $\gamma_{ji}$, computes the distance to the nearest target value in $\mathcal{S}$, encouraging convergence to simple, interpretable numbers.
        \item Regularization strength $\lambda \geq 0$: Controls the trade-off between prediction accuracy and interpretability.
    \end{itemize}
\end{enumerate}

\textbf{Constraints:}

\begin{enumerate}
    \item \textbf{Dimensional Homogeneity}: The constraint $\mathbf{D} \mathbf{w}_b^{(i)} = \mathbf{0}$ ensures that basis vectors $\mathbf{w}_b^{(i)}$ lie in the null space of the dimension matrix $\mathbf{D} \in \mathbb{R}^{r \times N}$, where $r$ is the number of fundamental dimensions (e.g., mass, length, time, temperature). This guarantees that combinations of variables using these weights are dimensionally consistent (dimensionless).
    
    \item \textbf{Null Space Constraint}: The requirement $\mathbf{w}_b^{(i)} \in \mathcal{N}(\mathbf{D})$ ensures that basis vectors are valid solutions to the dimensional homogeneity constraint, where $\mathcal{N}(\mathbf{D}) = \{\mathbf{w} \in \mathbb{R}^N : \mathbf{D}\mathbf{w} = \mathbf{0}\}$ is the null space of dimension $N-r$, with $r = \text{rank}(\mathbf{D})$.
    
    \item \textbf{Dominant Count Estimation}: The constraint $k = \text{estimate}(\mathbf{X}, \mathbf{y})$ indicates that the number of dominant dimensionless groups $k$ is estimated from the data using dimensionality reduction techniques (PCA or SIR) applied to the dimensionless variables $\mathbf{X}$ and output $\mathbf{y}$.
\end{enumerate}

\textbf{Problem Structure:}

The optimization problem searches over:
\begin{itemize}
    \item \textbf{Learnable parameters}: The coefficients $\{\boldsymbol{\gamma}_j\}_{j=1}^{k}$ and the neural network function $f$ (its weights and biases).
    \item \textbf{Fixed components}: The dimension matrix $\mathbf{D}$ (determined by physical units), basis vectors $\{\mathbf{w}_b^{(i)}\}$ (computed from $\mathbf{D}$ via dimensional analysis), and the number of groups $k$ (estimated from data).
\end{itemize}

The solution yields both the optimal dimensionless groups (through learned $\boldsymbol{\gamma}_j$ coefficients) and their functional relationship to the output (through learned $f$), providing a complete scaling law of the form $p^* = f(\Pi_1, \Pi_2, \ldots, \Pi_k)$.

To construct the dimensionless groups, the learnable coefficients $\boldsymbol{\gamma}_j$ linearly combine the basis vectors:
\begin{equation}
\mathbf{w}_j = \sum_{i=1}^{N-r} \gamma_{ji} \mathbf{w}_b^{(i)}, \quad \text{for } j = 1, \ldots, k
\label{eq:gamma_combination}
\end{equation}
where $\boldsymbol{\gamma}_j = [\gamma_{j1}, \gamma_{j2}, \ldots, \gamma_{j(N-r)}]^T$ are the learnable coefficients. Each dimensionless group is then computed as:
\begin{equation}
\Pi_j = \prod_{\ell=1}^{N} \tilde{p}_\ell^{w_{j\ell}}
\label{eq:dimensionless_group}
\end{equation}
where $w_{j\ell}$ is the $\ell$-th component of $\mathbf{w}_j$, and $\tilde{p}_\ell$ are the normalized input variables.

The following subsections provide pseudo-code for each module, followed by detailed mathematical descriptions.

\subsection{Data Preprocessing}

The data preprocessing module prepares the input data for dimensional analysis and optimization. Given raw experimental measurements or synthetic data, the module performs normalization and dimension matrix construction.

\subsubsection{Data Normalization}

Input variables $\mathbf{p} = [p_1, p_2, \ldots, p_N]$ are normalized to a common scale by dividing each variable by its maximum value:
\begin{equation}
\tilde{p}_i = \frac{p_i}{\max_{j=1,\ldots,M} p_{i,j}}
\label{eq:normalization}
\end{equation}
where $M$ is the number of data samples. This normalization ensures all variables are in the range $[0, 1]$ and prevents numerical issues during optimization. The output variable $p^*$ is normalized similarly.

\subsubsection{Dimension Matrix Construction}

For each input variable $p_i$, the module constructs a dimension vector $\mathbf{d}_i \in \mathbb{R}^d$ that encodes how the variable relates to $d$ fundamental dimensions (e.g., mass $M$, length $L$, time $T$, temperature $\Theta$). The dimension matrix $\mathbf{D} \in \mathbb{R}^{d \times N}$ is formed by stacking these vectors as columns:
\begin{equation}
\mathbf{D} = [\mathbf{d}_1, \mathbf{d}_2, \ldots, \mathbf{d}_N]
\label{eq:dimension_matrix}
\end{equation}
Each row of $\mathbf{D}$ corresponds to a fundamental dimension, and each column corresponds to an input variable. For example, if $p_i$ has dimensions $M^a L^b T^c \Theta^e$, then $\mathbf{d}_i = [a, b, c, e]^T$.

\subsection{Dimensional Analysis}

The dimensional analysis module computes the basis vectors that span the null space of the dimension matrix, providing the fundamental building blocks for constructing dimensionless groups.

\subsubsection{Null Space Computation}

According to the Buckingham $\pi$ theorem, if the rank of $\mathbf{D}$ is $r$, then there exist $N-r$ independent dimensionless groups. The basis vectors are computed as the null space of $\mathbf{D}$:
\begin{equation}
\mathcal{N}(\mathbf{D}) = \{\mathbf{w} \in \mathbb{R}^N : \mathbf{D}\mathbf{w} = \mathbf{0}\}
\label{eq:null_space}
\end{equation}

The null space is computed using singular value decomposition (SVD) or symbolic computation. If $\mathbf{D}$ has SVD $\mathbf{D} = \mathbf{U}\boldsymbol{\Sigma}\mathbf{V}^T$, then the columns of $\mathbf{V}$ corresponding to zero singular values form an orthonormal basis for the null space.

\subsubsection{Basis Vector Simplification}

The raw null space basis vectors are often not sparse or interpretable. The module simplifies them to primitive integer vectors for better interpretability. For each basis vector $\mathbf{v}$ in the null space:

\begin{enumerate}
    \item Convert to exact rational representation using symbolic computation.
    \item Clear denominators by multiplying by the least common multiple (LCM) of all denominators.
    \item Divide by the greatest common divisor (GCD) of all components to obtain a primitive integer vector.
\end{enumerate}

This yields basis vectors $\{\mathbf{w}_b^{(i)}\}_{i=1}^{N-r}$ with integer components that are easier to interpret physically.

\subsubsection{Basis Vector Normalization}

Optionally, the basis vectors can be normalized to unit length:
\begin{equation}
\mathbf{w}_b^{(i)} \leftarrow \frac{\mathbf{w}_b^{(i)}}{||\mathbf{w}_b^{(i)}||_2}
\label{eq:normalize_basis}
\end{equation}
where $||\cdot||_2$ denotes the Euclidean norm. Normalization ensures all basis vectors have unit magnitude, which can help with numerical stability during optimization.

\subsubsection{Dimensionless Variable Construction}

After obtaining basis vectors, dimensionless variables are constructed for each data sample:
\begin{equation}
\lg \Pi_{j,i} = \sum_{\ell=1}^{N} w_{b,j\ell}^{(i)} \lg \tilde{p}_{\ell,i}
\label{eq:lg_dimensionless}
\end{equation}
where $\lg$ denotes the logarithm (base 10 or natural), and $w_{b,j\ell}^{(i)}$ is the $\ell$-th component of the $i$-th basis vector. The logarithm transformation is used because dimensionless groups are products of powers, making the relationship linear in log-space.

\subsection{Dimensional Filtering}

The dimensional filtering module estimates the intrinsic dimensionality of the data, which determines the number of dominant dimensionless groups $k$ needed to represent the output. Two complementary methods are used: Principal Component Analysis (PCA) and Sliced Inverse Regression (SIR).

\subsubsection{Principal Component Analysis}

PCA identifies the dominant directions in the data by computing the eigendecomposition of the covariance matrix. The standardized data matrix $\mathbf{X} \in \mathbb{R}^{M \times (N-r)}$ (where each row is the logarithm of dimensionless variables for one sample) is centered and scaled:
\begin{equation}
\mathbf{X}_{\text{standardized}} = (\mathbf{X} - \boldsymbol{\mu}) \boldsymbol{\Sigma}^{-1}
\label{eq:pca_standardize}
\end{equation}
where $\boldsymbol{\mu}$ is the mean vector and $\boldsymbol{\Sigma}$ is a diagonal matrix of standard deviations.

The covariance matrix is computed as:
\begin{equation}
\mathbf{C} = \frac{1}{M-1} \mathbf{X}_{\text{standardized}}^T \mathbf{X}_{\text{standardized}}
\label{eq:pca_covariance}
\end{equation}

The eigendecomposition yields:
\begin{equation}
\mathbf{C} = \mathbf{V} \boldsymbol{\Lambda} \mathbf{V}^T
\label{eq:pca_eigen}
\end{equation}
where $\boldsymbol{\Lambda}$ is a diagonal matrix of eigenvalues $\lambda_1 \geq \lambda_2 \geq \cdots \geq \lambda_{N-r}$ and $\mathbf{V}$ contains the corresponding eigenvectors.

The explained variance ratio for the $i$-th principal component is:
\begin{equation}
r_i = \frac{\lambda_i}{\sum_{j=1}^{N-r} \lambda_j}
\label{eq:explained_variance}
\end{equation}

The number of dominant dimensionless groups $k$ is estimated as the smallest integer such that the cumulative explained variance exceeds a threshold (typically 0.75):
\begin{equation}
k = \min \left\{ j : \sum_{i=1}^{j} r_i \geq 0.75 \right\}
\label{eq:dominant_count_pca}
\end{equation}

\subsubsection{Sliced Inverse Regression}

SIR is a supervised dimensionality reduction method that uses output information to identify important directions, making it more robust for nonlinear relationships than PCA.

Given input data $\mathbf{X} \in \mathbb{R}^{M \times (N-r)}$ and output $\mathbf{y} \in \mathbb{R}^M$, SIR proceeds as follows:

\begin{enumerate}
    \item \textbf{Standardization}: Standardize inputs as in PCA (Equation~\ref{eq:pca_standardize}).
    
    \item \textbf{Slicing}: Partition the output into $H$ slices of approximately equal size:
    \begin{equation}
    S_h = \{i : y_{(h-1)M/H} \leq y_i < y_{hM/H}\}, \quad h = 1, \ldots, H
    \label{eq:sir_slices}
    \end{equation}
    where $y_{(j)}$ denotes the $j$-th order statistic of $\mathbf{y}$.
    
    \item \textbf{Slice Means}: Compute the mean of standardized inputs within each slice:
    \begin{equation}
    \boldsymbol{\mu}_h = \frac{1}{|S_h|} \sum_{i \in S_h} \mathbf{x}_i
    \label{eq:sir_slice_mean}
    \end{equation}
    where $\mathbf{x}_i$ is the $i$-th row of $\mathbf{X}_{\text{standardized}}$.
    
    \item \textbf{Covariance of Slice Means}: Compute the weighted covariance matrix:
    \begin{equation}
    \mathbf{C}_{\text{SIR}} = \sum_{h=1}^{H} \frac{|S_h|}{M} (\boldsymbol{\mu}_h - \bar{\boldsymbol{\mu}}) (\boldsymbol{\mu}_h - \bar{\boldsymbol{\mu}})^T
    \label{eq:sir_covariance}
    \end{equation}
    where $\bar{\boldsymbol{\mu}} = \sum_{h=1}^{H} (|S_h|/M) \boldsymbol{\mu}_h$ is the weighted mean of slice means.
    
    \item \textbf{Eigendecomposition}: Compute the eigendecomposition:
    \begin{equation}
    \mathbf{C}_{\text{SIR}} = \mathbf{V}_{\text{SIR}} \boldsymbol{\Lambda}_{\text{SIR}} \mathbf{V}_{\text{SIR}}^T
    \label{eq:sir_eigen}
    \end{equation}
    
    \item \textbf{Direction Selection}: The top $k$ eigenvectors corresponding to the largest eigenvalues indicate the most important directions for predicting the output.
\end{enumerate}

SIR is particularly effective when the relationship between dimensionless groups and output is nonlinear (e.g., exponential or logarithmic), as it uses output information to guide the search for relevant directions.

\subsection{Optimization and Discovery}

The optimization and discovery module searches for optimal coefficients $\{\boldsymbol{\gamma}_j\}_{j=1}^{k}$ and learns the functional relationship $f$ using a neural network architecture.

\subsubsection{Neural Network Architecture}

The neural network consists of a linear combination layer followed by hidden layers and an output layer. The architecture is designed to learn both the dimensionless group coefficients and the functional mapping.

\textbf{Linear Combination Layer}: The first layer computes linear combinations of the basis vectors:
\begin{equation}
\Pi_{j,i} = \sum_{\ell=1}^{N-r} \gamma_{j\ell} \lg \Pi_{b,\ell,i}
\label{eq:linear_combination}
\end{equation}
where $\lg \Pi_{b,\ell,i} = \sum_{m=1}^{N} w_{b,\ell m}^{(i)} \lg \tilde{p}_{m,i}$ is the logarithm of the dimensionless variable constructed from the $\ell$-th basis vector for sample $i$.

In matrix form, this layer has no bias term and applies no activation:
\begin{equation}
\mathbf{z}^{(1)} = \boldsymbol{\Gamma} \mathbf{X}
\label{eq:nn_linear_layer}
\end{equation}
where $\boldsymbol{\Gamma} \in \mathbb{R}^{k \times (N-r)}$ contains the learnable coefficients $\gamma_{ji}$, and $\mathbf{X} \in \mathbb{R}^{(N-r) \times M}$ contains the logarithm of basis dimensionless variables.

\textbf{Hidden Layers}: The hidden layers apply a nonlinear transformation:
\begin{equation}
\mathbf{z}^{(l+1)} = \text{ReLU}(\mathbf{W}^{(l)} \mathbf{z}^{(l)} + \mathbf{b}^{(l)}), \quad l = 1, \ldots, L-1
\label{eq:nn_hidden}
\end{equation}
where $\mathbf{W}^{(l)}$ and $\mathbf{b}^{(l)}$ are learnable weights and biases, and $\text{ReLU}(x) = \max(0, x)$ is the rectified linear unit activation function.

\textbf{Output Layer}: The final layer produces the prediction:
\begin{equation}
\hat{p}^* = \mathbf{W}^{(L)} \mathbf{z}^{(L)} + \mathbf{b}^{(L)}
\label{eq:nn_output}
\end{equation}

\subsubsection{Training Objective}

The network is trained to minimize the total loss (Equation~\ref{eq:complete_framework}). The prediction loss is the mean squared error between predicted and actual outputs:
\begin{equation}
\mathcal{L}_{\text{prediction}} = \frac{1}{M} \sum_{i=1}^{M} \left(p^*_i - \hat{p}^*_i\right)^2
\label{eq:nn_prediction_loss}
\end{equation}

The quantization regularization loss is computed as:
\begin{equation}
\mathcal{L}_{\text{quant}} = \sum_{j=1}^{k} \sum_{i=1}^{N-r} \min_{s \in \mathcal{S}} |\gamma_{ji} - s|
\label{eq:nn_quantization_loss}
\end{equation}
where $\mathcal{S}$ is the set of target values. Common choices for $\mathcal{S}$ include:
\begin{itemize}
    \item \textbf{Integers}: $\mathcal{S} = \{0, \pm 1, \pm 2, \pm 3, \pm 4, \pm 5\}$
    \item \textbf{Half-integers}: $\mathcal{S} = \{0, \pm 0.5, \pm 1, \pm 1.5, \pm 2, \pm 2.5, \pm 3\}$
    \item \textbf{Quarter-integers}: $\mathcal{S} = \{0, \pm 0.25, \pm 0.5, \pm 0.75, \pm 1, \pm 1.25, \ldots, \pm 3\}$
\end{itemize}

Note that zero is always included in $\mathcal{S}$ to encourage sparsity, allowing the model to exclude irrelevant basis vectors.

\subsubsection{Optimization Algorithm}

The network is trained using the Adam optimizer with learning rate $\alpha \in [10^{-4}, 10^{-3}]$. The training process involves:

\begin{enumerate}
    \item \textbf{Data Splitting}: Split the dataset into training and test sets (typically 80\% training, 20\% test).
    
    \item \textbf{Standardization}: Standardize inputs and outputs using mean and standard deviation:
    \begin{equation}
    \mathbf{X}_{\text{scaled}} = \frac{\mathbf{X} - \boldsymbol{\mu}_X}{\boldsymbol{\sigma}_X}, \quad \mathbf{y}_{\text{scaled}} = \frac{\mathbf{y} - \mu_y}{\sigma_y}
    \label{eq:nn_standardization}
    \end{equation}
    
    \item \textbf{Ensemble Training}: Train multiple models (typically 5-20) with different random initializations to capture solution diversity and improve robustness.
    
    \item \textbf{Iterative Updates}: For each epoch, compute gradients via backpropagation and update parameters:
    \begin{equation}
    \boldsymbol{\theta}^{(t+1)} = \boldsymbol{\theta}^{(t)} - \alpha \nabla_{\boldsymbol{\theta}} \mathcal{L}_{\text{total}}(\boldsymbol{\theta}^{(t)})
    \label{eq:adam_update}
    \end{equation}
    where $\boldsymbol{\theta}$ includes all network parameters $\{\boldsymbol{\Gamma}, \{\mathbf{W}^{(l)}, \mathbf{b}^{(l)}\}_{l=1}^{L}\}$.
\end{enumerate}